\documentclass[aps,pre,twocolumn,superscriptaddress]{revtex4-2}%
\usepackage{graphicx}
\usepackage{epsfig}
\usepackage{subfigure}  
\usepackage{float}
\usepackage{braket}
\usepackage{pslatex,latexsym}
\usepackage{bm,float,lipsum}
\usepackage{amsmath}
\usepackage{amssymb}
\usepackage{amsfonts}
\usepackage{mathrsfs}
\usepackage{color}
\begin{document}

\title{Predicting extreme events from data using deep machine learning: when and where} 

\author{Junjie Jiang}
\affiliation{The Key Laboratory of Biomedical Information Engineering of Ministry of Education, Institute of Health and Rehabilitation Science, School of Life Science and Technology, Research Center for Brain-inspired Intelligence, Xi'an Jiaotong University, The Key Laboratory of Neuro-informatics $\&$ Rehabilitation Engineering of Ministry of Civil Affairs, Xi'an, Shaanxi, China}
\affiliation{School of Electrical, Computer and Energy Engineering, Arizona State University, Tempe, Arizona 85287, USA}

\author{Zi-Gang Huang}
\affiliation{The Key Laboratory of Biomedical Information Engineering of Ministry of Education, Institute of Health and Rehabilitation Science, School of Life Science and Technology, Research Center for Brain-inspired Intelligence, Xi'an Jiaotong University, The Key Laboratory of Neuro-informatics $\&$ Rehabilitation Engineering of Ministry of Civil Affairs, Xi'an, Shaanxi, China}

\author{Celso Grebogi}
\affiliation{Institute for Complex Systems and Mathematical Biology, School of Natural and Computing Sciences, King's College, University of Aberdeen, AB24 3UE, UK}

\author{Ying-Cheng Lai} \email{Ying-Cheng.Lai@asu.edu}
\affiliation{School of Electrical, Computer and Energy Engineering, Arizona State University, Tempe, Arizona 85287, USA}
\affiliation{Department of Physics, Arizona State University, Tempe, Arizona 85287, USA}

\date{\today}

\begin{abstract}

We develop a deep convolutional neural network (DCNN) based framework for model-free prediction of the occurrence of extreme events both in time (``when'') and in space (``where'') in nonlinear physical systems of spatial dimension two. The measurements or data are a set of two-dimensional snapshots or images. For a desired time horizon of prediction, a proper labeling scheme can be designated to enable successful training of the DCNN and subsequent prediction of extreme events in time. Given that an extreme event has been predicted to occur within the time horizon, a space-based labeling scheme can be applied to predict, within certain resolution, the location at which the event will occur. We use synthetic data from the 2D complex Ginzburg-Landau equation and empirical wind speed data of the North Atlantic ocean to demonstrate and validate our machine-learning based prediction framework. The trade-offs among the prediction horizon, spatial resolution, and accuracy are illustrated, and the detrimental effect of spatially biased occurrence of extreme event on prediction accuracy is discussed. The deep learning framework is viable for predicting extreme events in the real world.   

\end{abstract}

\maketitle

\section{Introduction}

One of the outstanding and most challenging problems in nonlinear dynamics
and complex systems is predicting extreme events in spatiotemporal chaotic 
systems. Extreme events, also known as rare and intense events, occur in a 
variety of physical systems~\cite{AJK:book}. Familiar examples 
include earthquakes, intense tropical cyclones, tornadoes, and rogue 
ocean waves. For example, rough waves in the ocean can arise from long-range
acoustic wave propagation through ocean's sound channel~\cite{WF:1998}.
In optics (e.g., optical fiber systems), extreme events can manifest themselves
as waves with an abnormally large amplitude - rogue waves~\cite{SRKB:2007,
Akhmedievetal:2016,SCLSBCB:2016}. Extreme events can also occur in
engineering systems, e.g., complex networked or online systems, examples
of which range from sudden bursts of packet flows on the internet, jamming 
in computer or transportation networks, denial-of-service cyberattacks, and 
blackouts due to load imbalances in the electrical power grid.
There have been efforts in statistical characterization of extreme events 
and in understanding their dynamical origin~\cite{AJK:book,LPP:2001,
SK:2005,AK:2005,NBN:2006,SK:2008,AKLF:2013,
BKT:2013,KAFL:2014,MBK:2015,QM:2016,MS:2018,BSM:2019}. For example, the 
statistical distribution~\cite{LPP:2001} and the recurrent or long-term 
correlation property~\cite{AK:2005,SK:2008} of extreme events were studied. 
There were also efforts in predictability~\cite{HAHK:2007,DB:2008,
COSOG:2013,BF:2017} and model-based control~\cite{NO:2007,DCLX:2008,
BAK:2015} of extreme events. Catastrophic 
events in the past two decades include the crash of China Airlines Flight 611 
due to metal fatigue in 2002, the Fukushima Daiichi nuclear plant meltdown 
due to a tsunami wave triggering boiling water reactors shutdown in 2011, and 
hurricane Harvey in the Gulf of Mexico in 2017 that took away the lives of
more than 100 people and caused \$125 billion damage~\cite{Lopatka:2019}. 
Due to global-warming induced climate 
change~\cite{HSRLO:2000}, extreme events are likely to occur at an 
increasingly high frequency with severe social, environmental and economical 
damages~\cite{EEGKKA:2000,PR:2002,CC:2012,NOAA:2016,Stott:2016}. 
While there have been recent efforts in model based prediction of extreme
events~\cite{QM:2018,MMQ:2019}, in realistic situations, an accurate 
model underlying the dynamical process leading to an extreme event often is 
not available. It is imperative to develop model-free prediction methods 
based solely on data. This problem, despite its paramount importance, has
remained outstanding.

In this paper, we develop a deep machine learning framework to predict extreme 
events without requiring any model of the underlying system. Recent years have 
witnessed tremendous development in machine learning, especially various 
deep learning methods~\cite{LBH:2015,JM:2015,GBC:book,HZRS:2016} based on 
artificial neural networks, with applications in a wide variety of areas. 
For example, deep learning has been used to recognize speech and visual 
objects~\cite{KSH:2012,HDYDMJSVNK:2012,HZRS:2016}, to predict the potential 
structure-activity relationships among drug molecules~\cite{MSLDS:2015}, 
to detect major depressive disorder from EEG signals~\cite{DGSLCG:2020}, and
even to beat the human player in Go game~\cite{SSSAHGHBLB:2017}. On predicting 
severe weather and climate events, methods based on artificial neural networks 
or artificial intelligence have been deemed important and potentially 
feasible~\cite{Lopatka:2019,RCSJDC:2019}. For example, deep convolutional 
neural networks (DCNNs) have been used to predict the EI Ni\~no/South 
Oscillations~\cite{HKL:2019} that occur over a much longer time scale (e.g., 
months or a year) than those of extreme weather events such as intense tropical 
cyclones (e.g., days or weeks). A binary classification scheme was formulated 
for deriving optimal predictors of extremes directly from data~\cite{GS:2019}. 
More recently, a densely connected mixed-scale network model was 
proposed~\cite{QM:2020} to predict extreme events in the truncated 
Korteweg–de Vries equation - a nonlinear partial differential equation (PDE) 
in one spatial dimension which describes the complex dynamics in surface water 
wave turbulence. Our focus here is on DCNN based prediction of extreme events 
in terms of the ``when'' and ``where'' issues using synthetic data from 
spatiotemporal nonlinear dynamical systems and empirical image data from a 
real atmospheric system. 

We consider the setting of a two-dimensional (2D) closed domain in which 
extreme events occur. This setting is quite ubiquitous for studying extreme 
events in the natural world, such as earthquakes that occur in a finite 
geophysical area or extreme wind speed in an oceanic region. Physically, the 
extreme events represent exceptionally high amplitude or intensity of some 
``field'' such as the seismological wave or the velocity field of the wind 
across a specific spatial region. For simplicity, we consider a square 2D 
domain in which extreme events occur randomly in both space and time. If we 
color-code the amplitude or intensity of the field, its distribution at any 
instant of time is effectively an image - a snapshot. An extreme event at 
some instant of time will manifest itself as a bright spot somewhere in the 
image. As the system evolves in time, a large number of snapshots of the 
field distribution in the spatial domain can be taken. Some of these snapshots 
or images may contain an extreme event, while many others may not. Properly 
labeled images serve naturally as inputs to a DCNN for training and prediction.

We exploit DCNN to predict extreme events, focusing on the critical questions
of ``when'' and ``where,'' i.e., when and where in the region of interest do 
extreme events occur? To address the ``when'' question, we distinguish or 
label the instantaneous state distribution of the system (the images) into 
two classes: one in an appropriate time interval preceding the occurrence 
of an extreme event (e.g., labeled as {\bf 1}) and another with no extreme 
event in the next time interval of the same length (labeled as {\bf 0}). We 
then train a DCNN with a large number of randomly mixed labeled images. After 
successful completion of training, the DCNN possesses the predictive power 
in that, when a new image is presented to it, the deep networked system can 
output the correct label with reasonable accuracy, effectively solving the 
``when'' problem by predicting whether an extreme event is going to occur in 
the next time interval. The initially pre-defined time interval is 
thus the time horizon of prediction. Given that an extreme event has been 
predicted to occur within the horizon, we can address the ``where'' question 
by focusing on the images that carry a {\bf 1} label. In particular, we divide 
the spatial domain into a grid of cells of equal area. For example, for a 
$4\times 4$ grid, each cell carries a label, e.g., from {\bf 0} to {\bf 15}.
If the extreme event occurs in one of the cells, e.g., cell 12, then this 
image carries the label {\bf 12}, and so on. This way, the available images 
are partitioned into 16 classes, and the DCNN can be trained with a large 
number of such labeled images for gaining the power to predict in which 
spatial cell would an extreme event occur, effectively solving the ``where'' 
problem. The size of each cell determines the spatial resolution of 
prediction. There are trade-offs among the three key indicators: the time 
horizon, the spatial resolution, and the prediction accuracy. In general,
increasing the prediction horizon or the spatial resolution or both will 
reduce the accuracy.

Taken together, our DCNN based prediction paradigm consists of two steps. 
Firstly, we predict whether an extreme event would occur in a given time 
interval in the future (the ``when'' problem). Secondly, given that
an extreme event is going to occur within the time horizon, we predict 
the location of the event with certain spatial resolution (the ``where'' 
problem). For illustration, we use synthetic data from a model and empirical
data from a real atmospheric system: the 2D complex Ginzburg-Landau equation 
(CGLE) and the wind speed distribution in the North Atlantic ocean. In each 
case, we quantify the prediction results, demonstrate that reasonably high 
prediction accuracy can be achieved, and discuss the trade-offs among the 
time horizon, spatial resolution, and prediction accuracy. 

\section{Results}

We choose two classic and widely used DCNNs to predict extreme events in 
spatiotemporal dynamical systems: LeNet-5~\cite{LBBH:1998} and 
ResNet-50~\cite{HZRS:2016}. LeNet-5 was first articulated for classification 
of handwritten characters in 1998, and it has inspired the development of
state-of-the-art DCNNs for tasks such as image recognition~\cite{KSH:2012}, 
image segmentation~\cite{CGGS:2012}, and object detection~\cite{FCNL:2012}. 
The DCNN ResNet-50, a much ``deeper'' neural network in that it
has substantially more hidden layers than LeNet-5, was developed in 2016, 
where a residual learning framework was introduced to expedite training
(ResNet was the winner of the ILSVRC 2015 classification task). We shall 
demonstrate that both LeNet-5 and ResNet-50 have the power to predict extreme
events.

\begin{figure*}[ht!]
\centering
\includegraphics[width=\linewidth]{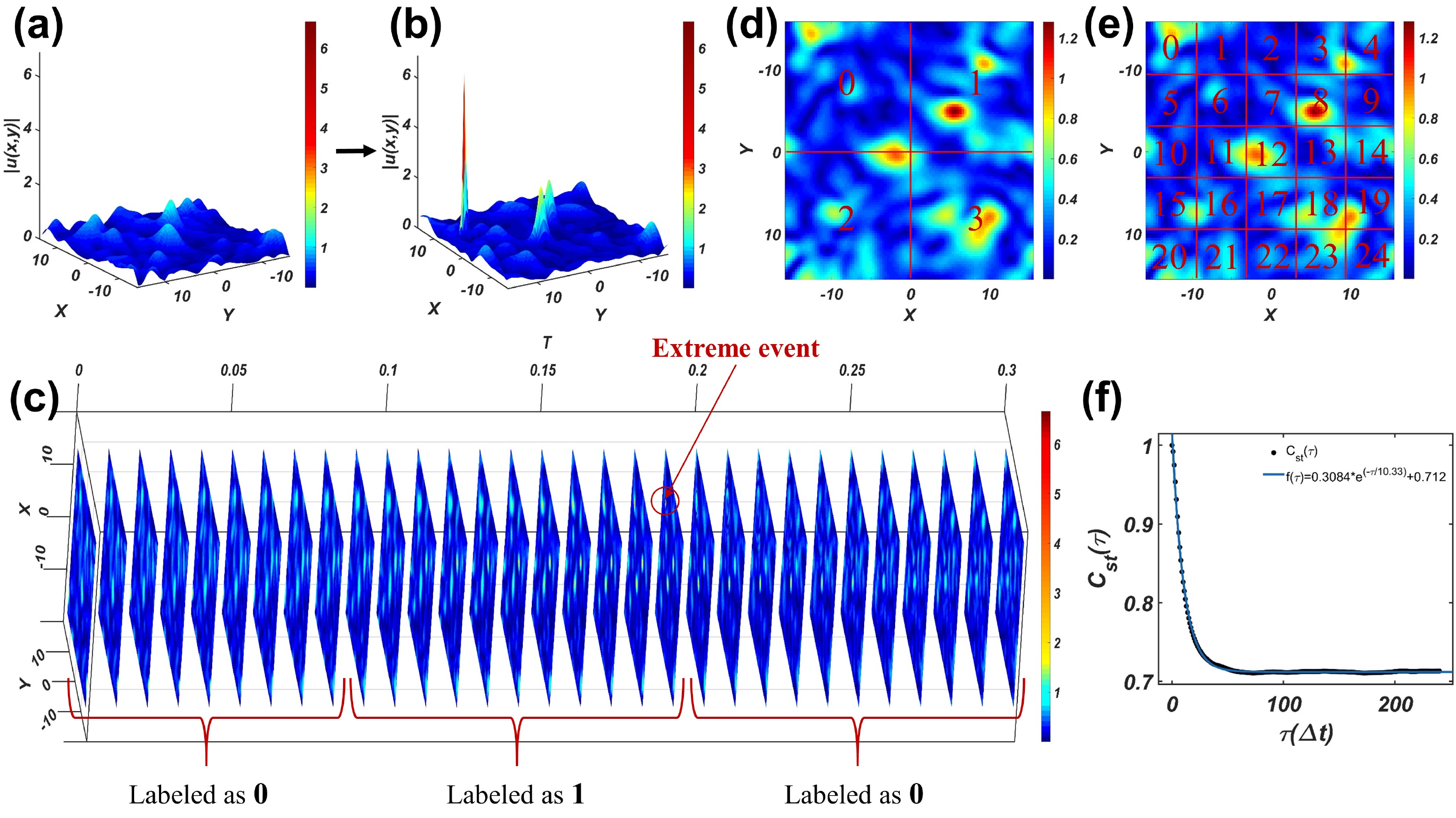}
\caption{ Representative 2D states of CGLE without or with an extreme 
event and an example of training data labeling scheme for DCNN.
(a) A 2D state without any extreme event and (b) a state with an extreme event
that occurs ten time steps after the state in (a), where an extreme event is 
defined if the maximum magnitude of the instantaneous amplitude in space 
exceeds the threshold value $|\bar{u}| = 5$. (c) A sequence of 31 
snapshots that occur consecutively in time (total time duration $=0.3$), 
separated by time step $\Delta t$, together with a labeling scheme for 
predicting ``when'' an extreme event will occur, where the state with an 
extreme event and all preceding $p=10$ states are labeled as ${\bf 1}$, and 
all other states are labeled as ${\bf 0}$. The time horizon of prediction is
$p\Delta t$. In general, there is a trade-off between the prediction horizon
and accuracy. (d) A labeling scheme for predicting the location of the 
occurrence of the extreme event, where the entire domain is divided into 
a $2\times 2$ grid with the four cells labeled as ${\bf 0}$, ${\bf 1}$, 
${\bf 2}$, and ${\bf 3}$, respectively. The color coding is based on the 
magnitude $|u|$ of the complex field. (e) A finer labeling scheme by which 
the domain is divided into a $5\times 5$ grid, generating 25 classes of 
images. There is a trade-off between spatial resolution and prediction 
accuracy as well. (f) Temporal correlation function $C_{st}(\tau)$ of 2D CGLE 
system, where the black dots and the blue curve are the actual correlation 
function and the exponential fitting. The time constant $\tau_c$ is about 
$10.33$. (See Appendix~\ref{appendix_A} for the definition of $C_{st}(\tau)$ 
and the simulation setting.)}
\label{fig:2DCGL_P}
\end{figure*}

\subsection{Predicting extreme events based on synthetic data from a 
paradigmatic model of nonlinear physical systems}

To test and demonstrate the power of DCNNs to predict extreme events in a 
controlled setting, we exploit a general mathematical model of diverse 
physical systems to generate synthetic datasets. In particular, we consider 
a finite 2D domain in which a set of physical variables evolve according to 
nonlinear dynamical rules and extreme events occur infrequently at different 
locations from time to time. From a dynamical systems point of view, extreme 
events are the consequence of the nonlinear interactions among different 
components of the system and are typically due to constructive interference 
among these components. For example, in a wave system, the wave amplitude at 
a spatial location at any time is the result of the superposition of many 
wave packets at different spatial locations from some earlier time. If 
the phases of most wave packets become coherent at certain time instant, a 
large amplitude event can arise at this moment. Phase coherence, however, is 
rare, and so are extreme wave events. Mathematically, it is convenient to 
describe the system by nonlinear PDEs in spatial dimension two. Because of 
nonlinearity, sensitive dependence on initial conditions, parameter 
fluctuations or external perturbations can arise. As a result, extreme events 
can occur throughout the domain at random time and locations. The general 
setting describes real-world phenomena such as material failures for which 
two-dimensional images can be obtained or earthquakes in a geographical region.

To be concrete, we study a closed PDE system of spatial dimension two as 
described by the CGLE (Appendix~\ref{appendix_A}), which is a paradigmatic 
model for gaining insights into a variety of physical phenomena such as 
nonlinear waves in optical fibers, chemical reactions, superconductivity, 
superfluidity, Bose-Einstein condensation, and liquid 
crystals~\cite{AK:2002,CH:1993,Kbook:1984}. Figures~\ref{fig:2DCGL_P}(a) and 
\ref{fig:2DCGL_P}(b) show two representative snapshots of the amplitude
distribution of the complex field $u(x,y)$ at two instants of time, one 
without and another with an extreme event as indicated by the localized, 
high-amplitude peak, respectively, where the former occurs ten time steps 
before the latter. A close examination of the 2D state in 
Fig.~\ref{fig:2DCGL_P}(a) gives no indication that an extreme event is going 
to occur in the near future, let alone its spatial location. 

To enable a DCNN to predict extreme events in both time and space, an essential
step is to articulate proper labeling schemes for the 2D states. 
Figure~\ref{fig:2DCGL_P}(c) illustrates a labeling scheme for predicting when
an extreme event will occur, where all prior 2D states within $p$ time steps 
of the occurrence of an extreme event (including the 2D state containing the 
event) are labeled as ${\bf 1}$, and all other 2D states are labeled as 
${\bf 0}$. 
In general, the value of $p$ determines the prediction horizon and accuracy,
and the trade-off between them provides a criterion for choosing $p$. To see
this, it is convenient to use the mean temporal period or the time constant
$\tau_c$ underlying the temporal correlation function of the state evolution
as a reference time scale. For $p < \tau_c$, the prediction horizon is short,
but the 2D states labeled as ${\bf 1}$ are highly correlated with the state
containing the extreme event, making it possible to achieve a high prediction
accuracy. However, for $p > \tau_c$, the prediction horizon is long but the 
accuracy will be sacrificed. Empirically, a proper choice of $p$ is
$p \approx \tau_c$. In Fig.~\ref{fig:2DCGL_P}(c), we have $p = 10$. 

Given that an extreme event has been predicted to occur within $p$ time steps,
we can predict its location by dividing the entire 2D domain into a uniform 
grid with a distinct label for each cell. For example, 
in Fig.~\ref{fig:2DCGL_P}(d) where the 2D state is the same as that in 
Fig.~\ref{fig:2DCGL_P}(a), the spatial domain is divided into a $2\times 2$
grid with four cells labeled as ${\bf 0}$, ${\bf 1}$, ${\bf 2}$, and ${\bf 3}$,
respectively, where a specific label indicates the cell in which the extreme
event will occur. Alternatively, a higher spatial resolution can be used, as
shown in Fig.~\ref{fig:2DCGL_P}(e), where the domain is divided into a 
$5\times 5$ grid. As we will demonstrate, there is a trade-off between spatial
resolution and prediction accuracy: a higher resolution generally reduces the
accuracy. 
 
\begin{figure}[ht!]
\centering
\includegraphics[width=\linewidth]{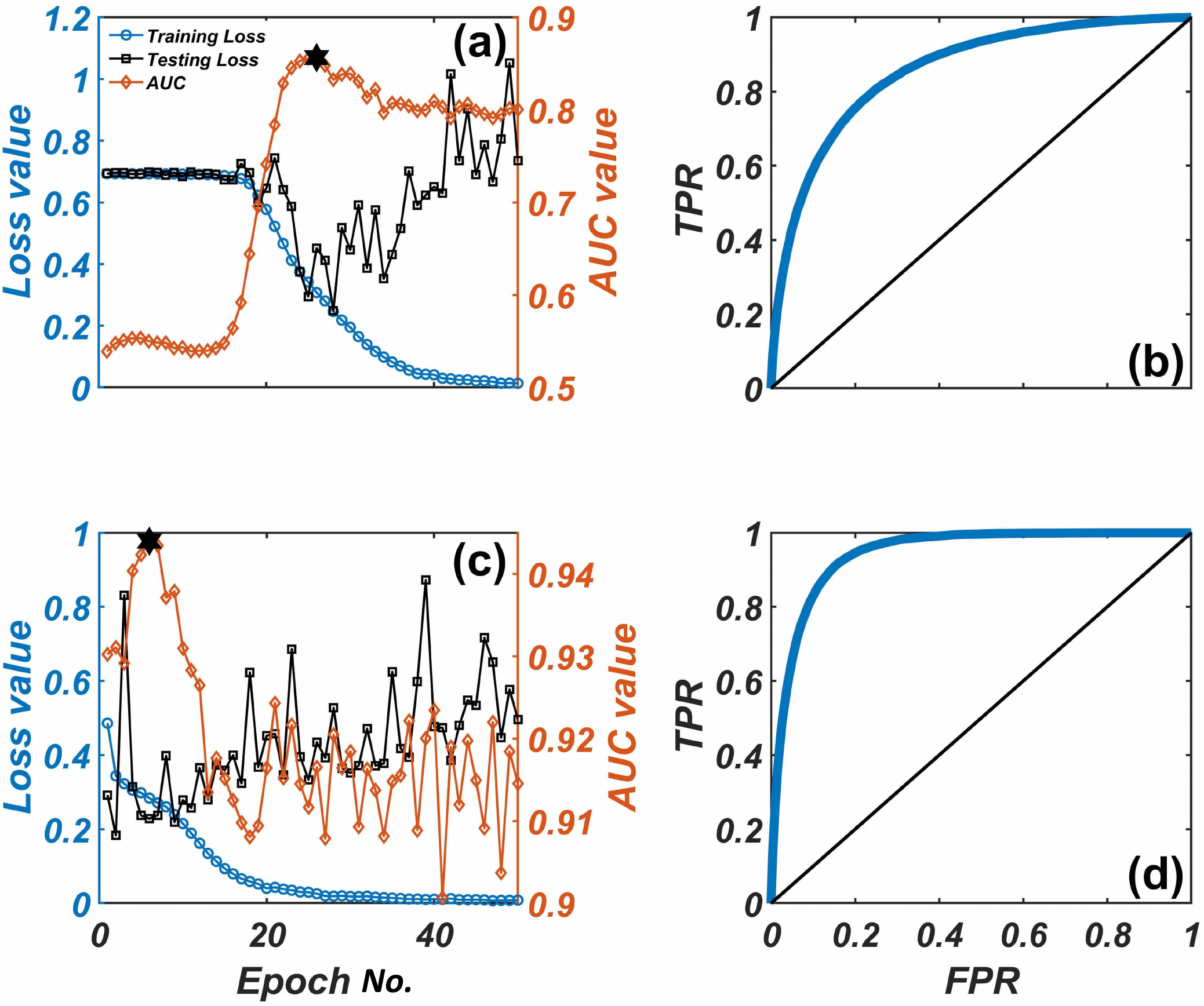}
\caption{Predicting occurrence of an extreme event with DCNN - the issue of 
``when''. Two types of DCNNs are used: LeNet-5 and ResNet-50. (a) With LeNet-5,
training loss (blue circles), testing loss (black squares), and AUC value 
(brown diamonds) versus the number of training epochs.
The black hexagram marker represents the epoch at which the best AUC value is 
achieved. (b) The ROC curve corresponding to the black hexagram in (a). 
(c) Results similar to those in (a) but from ResNet-50, with the same legends 
as in (a). (d) The ROC curve corresponding to the black hexagram in (c). The 
amplitude threshold for defining an extreme event is $|\bar{u}| = 5$.}
\label{fig:2DCGL_when}
\end{figure}

We now address the ``when'' question, i.e., to predict the possible occurrence
of an extreme event within a reasonable time interval in the future. Because of 
our labeling scheme to associate an extreme event with $p$ snapshots of 2D 
states prior to or at its occurrence, the prediction horizon is $p$ time steps,
i.e., $p\Delta t$, where $\Delta t$ is the sampling time interval. Ideally,
after training is completed, the DCNN should be able to give a label, either 
${\bf 0}$ or ${\bf 1}$, to any snapshot of the system state in space during
its course of evolution. At a given time with a specific snapshot as input, 
if the DCNN outputs label ${\bf 1}$, an extreme event is predicted to occur 
within the next $p$ time steps, with uncertainty determined by the prediction 
accuracy. If the output label is deemed to be ${\bf 0}$, there will be no 
extreme event within the prediction horizon counted from the present time.

To quantify the performance of the DCNN for predicting when an extreme event
would occur, we use the receiver operating characteristic (ROC). A ROC curve   
is generated with the true-positive rate (TPR) and the false-positive rate 
(FPR) on the Y and X-axis, respectively. The best possible performance 
corresponds to the top left corner (TPR = 1 and FPR = 0), so the area under 
the ROC curve (AUC) is one - the maximally possible value. In general, large
AUC values indicate a better performance. (A detailed description of how the 
ROC curve is calculated can be found in Appendix~\ref{appendix_B}.)

Figure~\ref{fig:2DCGL_when} presents a representative example for predicting
extreme events in the CGLE, where one training dataset has $100007$ 2D states, 
out of which $50006$ and $50001$ 2D states are labeled as ${\bf 1}$ and 
${\bf 0}$, respectively. A larger dataset is also tested, which contains
$224010$ 2D states with $15800$ and $208210$ states labeled as ${\bf 1}$ and 
${\bf 0}$, respectively. (A detailed description of how the training datasets 
are generated can be found in Appendix~\ref{appendix_A}.) 
Figures~\ref{fig:2DCGL_when}(a) and \ref{fig:2DCGL_when}(c) show the loss 
values and the AUC for training and testing with LeNet-5 and 
ResNet-50, respectively. Both DCNNs are trained using the deep learning 
framework Pytorch~\cite{PGCCYDLDAL:2017,PGMLBCKLGA:2019}, with binary 
cross-entropy loss function and a stochastic gradient descent optimizer. 
The relevant parameters are: batch size 128, learning rate $10^{-3}$,
momentum factor 0.9, and L2 penalty $5\times10^{-4}$ (see description in SI - 
Supporting Information~\cite{SI}). 
From Fig.~\ref{fig:2DCGL_when}(a), we see that, for
LeNet-5, the training loss starts to decrease steadily at about the $20$th 
training epoch, after which the training loss keeps decreasing as the number 
of training epochs is increased, but the testing loss decreases initially 
and then starts to increase slowly. For the test dataset, the best AUC value 
(about 0.857, marked as a black hexagram) is achieved at about the $26$th 
epoch. Figure~\ref{fig:2DCGL_when}(b) shows the corresponding ROC 
curve for LeNet-5. While the ROC curve involves 
a series of systematically varying threshold values for distinguishing the two 
output labels, a natural choice is $0.5$, i.e., an output value larger or 
smaller than $0.5$ represents prediction of class ${\bf 1}$ or ${\bf 0}$, 
respectively. For this choice of the threshold, the numbers of true-positive, 
false-positive, false-negative, and true-negative cases are $N_{TP}=11501$, 
$N_{FP} = 37141$, $N_{FN} = 4299$ and $N_{TN} = 171069$, respectively, 
so the values of $\mbox{TPR}\equiv N_{TP}/(N_{TP}+N_{FN})$ and 
$\mbox{FPR}\equiv N_{FP}/(N_{FP}+N_{TN})$ are $72.79\%$ and $17.84\%$, 
respectively. For LeNet-5, the overall prediction accuracy is 
approximately $81.5\%$. The corresponding results for ResNet-50 are shown in 
Figs.~\ref{fig:2DCGL_when}(c) and \ref{fig:2DCGL_when}(d). We see that only 
six training epochs are necessary to achieve the best performance with 
$\mbox{AUC}\approx 94.4\%$ (the prediction accuracy), after which the 
training loss decreases gradually as the number of training epochs is 
increased. However, there are fluctuations associated with the testing loss. 
The ROC curve corresponding to the best AUC value is shown in 
Fig~\ref{fig:2DCGL_when}(d). For the threshold value $0.5$, the numbers of 
true-positive, false-positive, false-negative, and true-negative cases are 
$12930$, $19742$, $2870$, and $188468$, respectively, giving the values of 
TPR and FPR as $81.84\%$ and $9.48\%$, respectively. 

\begin{figure*}[ht!]
\centering
\includegraphics[width=0.9\linewidth]{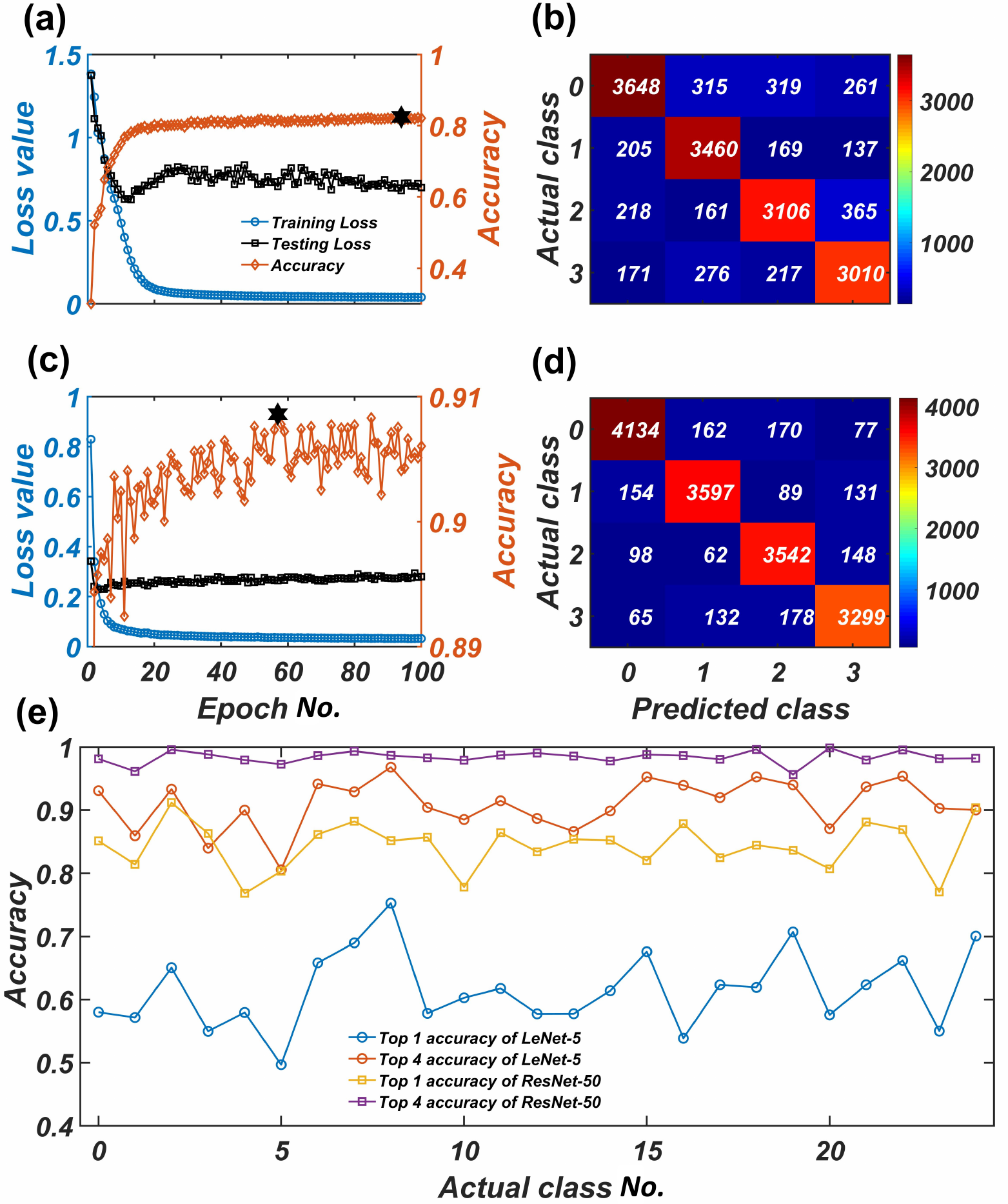}
\caption{Predicting the occurrence of an extreme event with DCNN - the issue 
of ``where''. Given that an extreme event has been predicted to occur within
$p$ time steps in the future, a labeling scheme based on a uniform spatial 
grid can be used to train the DCNNs. (a) For LeNet-5, training loss (blue 
circles), testing loss (black squares), and accuracy (brown diamonds) versus 
the length of the training phase, where the black hexagram marker represents 
the training epoch at which the best accuracy is achieved. (b) The confusion 
matrix associated with the black hexagram in (a). (c,d) Results corresponding 
to those in (a,b), respectively, but for ResNet-50. In (a-d), the spatial 
domain is divided into a $2\times2$ grid. (e) For both LeNet-5 and
ResNet-50 with a $5\times 5$ division of the domain, the top-one (one cell) 
and top-four (four cells with the highest accuracy values combined) accuracy 
for each class for predicting where the extreme will occur in the domain.
For ResNet-50, the top-four accuracy is better than $95\%$. The amplitude 
threshold for defining an extreme event is $|\bar{u}| = 5$.} 
\label{fig:2DCGL_where}
\end{figure*}

Having used DCNN to predict that an extreme event will occur in the next
$p$ time steps, it is necessary to predict where it will occur in the spatial
domain of interest. Here we demonstrate that this ``where'' issue can be 
addressed using DCNNs as well. Specifically, we use an independent DCNN and 
carry out the training beforehand, as follows. We gather all the state 
snapshots carrying the label ${\bf 1}$ from the original dataset. Take the 
case in Fig.~\ref{fig:2DCGL_when} as an example. The original dataset contains 
$50006$ such 2D states. If we divide the space into four identical subcells 
with the labels ${\bf 0}$, ${\bf 1}$, ${\bf 2}$, and ${\bf 3}$, the respective 
numbers of snapshots are $12628$, $12870$, $12320$, and $12188$. Note that 
these numbers are approximately the same, indicating that the dataset is 
{\em unbiased} in the space in the sense that the probability for an extreme
even to occur is roughly uniform across the entire domain. The whole dataset
is divided into a training dataset of $33968$ and a testing dataset of $16038$ 
snapshots. In the testing set, there are $4543$, $3971$, $3850$, and $3674$ 2D 
states that are labeled as class ${\bf 0}$, ${\bf 1}$, ${\bf 2}$, and ${\bf 3}$,
respectively. Figures~\ref{fig:2DCGL_where}(a) and \ref{fig:2DCGL_where}(c) 
show the loss and accuracy values during the training of DCNN LeNet-5 and 
ResNet-50, respectively, where the accuracy is calculated based on the whole 
testing dataset, i.e., with respect to all four classes. As for the case of 
predicting when an extreme event will occur, both DCNNs are trained using the 
deep learning framework Pytorch~\cite{PGCCYDLDAL:2017,PGMLBCKLGA:2019} with 
the same training parameter values as those in Fig.~\ref{fig:2DCGL_when}. At 
the beginning of the training phase, the values of testing loss and accuracy 
change quite rapidly. However, after several training epochs, both types of 
DCNNs have attained stable testing loss and accuracy. For LeNet-5 and 
ResNet-50, optimal training with the highest accuracy is achieved at 
epoch $94$ and $57$, respectively. (Note that the best accuracy does not 
correspond to the lowest testing loss.) 
Figures~\ref{fig:2DCGL_where}(b) and \ref{fig:2DCGL_where}(d) 
present the confusion matrices of the predicted class versus the actual class 
for the testing result of the best accuracy for LeNet-5 and ResNet-50, 
respectively, where the values of the diagonal elements indicate correct 
prediction while those of off-diagonal elements represent errors. We see that
the diagonal elements are much larger than the off-diagonal elements, 
suggesting the effectiveness of prediction. For LeNet-5, the prediction 
accuracy for classes ${\bf 0}$, ${\bf 1}$, ${\bf 2}$, and ${\bf 3}$ are
approximately $80.30\%$, $87.13\%$, $80.68\%$ and $81.93\%$, respectively, 
while the respective accuracy values for ResNet-50 are higher: $91.00\%$, 
$90.58\%$, $92.00\%$, and $89.79\%$. 

We have also tested a spatial grid of higher resolution: $5\times 5$. In this
case, there are 25 labels, corresponding to 25 training classes. The 
prediction accuracy for all classes is shown in Fig.~\ref{fig:2DCGL_where}(e)
for both LeNet-5 and ResNet-50. During prediction, for any input image, the 
output of the DCNN is a $25\times 1$ vector, each element of which gives the 
respective probability for each predicted class. Two types of accuracy are 
calculated: top-1 accuracy, which is simply the prediction accuracy for each 
class, and top-4 accuracy, which for each class is defined as the accuracy 
of predicting the occurrence of the extreme event in the four cells with the 
highest predicted probability values. If such four cells contain the actual 
class, the top-four accuracy would be $100\%$; otherwise, it is zero. 
Averaging over all images of the specific class gives the top-four accuracy 
value, as shown in Fig.~\ref{fig:2DCGL_where}(e). We have that the average 
values of the top-one accuracy for LeNet-5 and ResNet-50 are approximately 
$61.49\%$ and $84.34\%$, respectively, indicating that ResNet-50 is more 
effective at predicting the location of occurrence of the extreme event. 
The mean top-4 accuracy achieved with ResNet-50 is remarkably high: $98.37\%$. 
(The corresponding confusion matrix is presented in SI~\cite{SI}.)

\begin{figure*}[ht!]
\centering
\includegraphics[width=\linewidth]{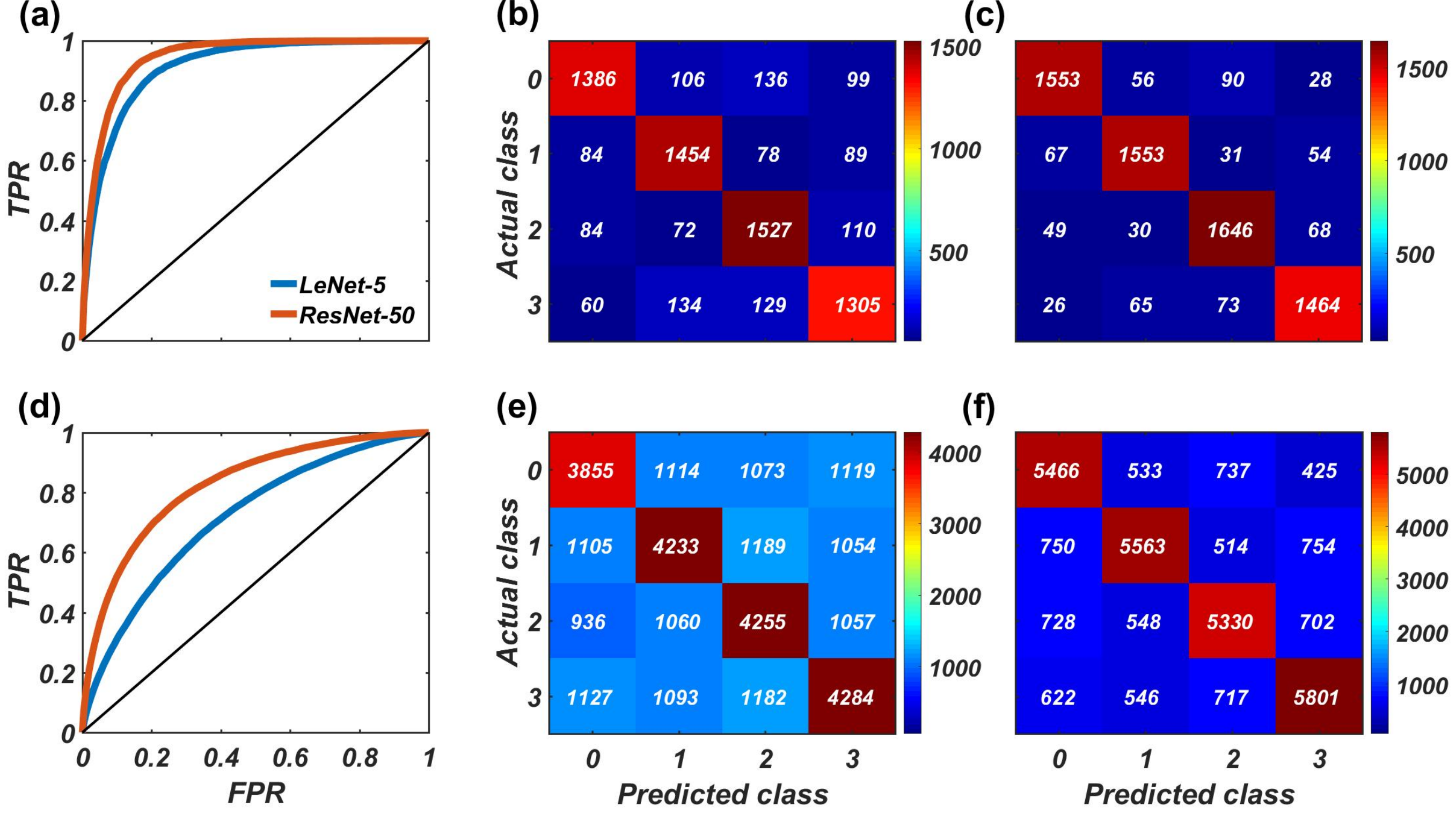}
\caption{Performance robustness of DCNN based prediction of extreme events in 
terms of occurrence time and spatial location in CGLE. 
(a-c) ROC curves and confusion matrices for the same dataset as in 
Figs.~\ref{fig:2DCGL_when} and \ref{fig:2DCGL_where} but with the threshold
$|\bar{u}|$ in the amplitude for defining an extreme event increased from 5 
to 6, for $p=10$. (d-f) ROC curves and confusion matrices for $|\bar{u}| = 5$ 
and $p=20$. In (a) and (d), the best ROC curves for LeNet-5 and ResNet-50
are presented, respectively. Panel pairs (b,e) and (c, f) show the 
corresponding confusion matrices for LeNet-5 and ResNet-50, respectively. }
\label{fig:2DCGL_D_th_LT}
\end{figure*} 

To demonstrate the robustness of DCNN based prediction of extreme events, 
we study the performance characteristics for different values of the 
amplitude threshold $|\bar{u}|$ for defining an extreme event. 
Figure~\ref{fig:2DCGL_D_th_LT}(a) show, for the same data set
as in Figs.~\ref{fig:2DCGL_when} and \ref{fig:2DCGL_where}, results for 
predicting ``when'' an extreme event will occur for $|u| = 6$. Because of 
the higher amplitude threshold value, the number of states labeled as ${\bf 1}$ 
is smaller: in this case there are $6780$ and $217230$ 2D states labeled as 
${\bf 1}$ and ${\bf 0}$ in the dataset, respectively. What is displayed
in Fig.~\ref{fig:2DCGL_D_th_LT}(a) is the best ROC curves for LeNet-5 and 
ResNet-50, where the corresponding AUC values are approximately $0.91$ and 
$0.94$, respectively. Comparing Fig.~\ref{fig:2DCGL_D_th_LT}(a) with 
Figs.~\ref{fig:2DCGL_when}(b) and \ref{fig:2DCGL_when}(d), we see that a
small increase in the amplitude threshold value improves the performance of 
LeNet-5 but that of ResNet-50 is largely unaffected. The fact that the AUC 
values are above 0.9 indicates robust predictability of both types of DCNNs
for extreme events. Figure~\ref{fig:2DCGL_D_th_LT}(b) shows the corresponding
confusion matrix for LeNet-5 for predicting ``where'' the extreme event, 
which has been predicted to occur within the time horizon [as characterized
by Fig.~\ref{fig:2DCGL_D_th_LT}(a)], will occur, where the spatial domain is 
divided into a $2\times 2$ grid with four spatial labels (classes) to be 
predicted. The prediction accuracies for the extreme event to occur in the 
four subdomains (classes ${\bf 0}$ to ${\bf 3}$) are approximately $80.25\%$, 
$85.28\%$, $85.16\%$, and $80.16\%$, respectively. The corresponding results 
for ResNet-50 are shown in Fig.~\ref{fig:2DCGL_D_th_LT}(c), where the 
prediction accuracies for classes ${\bf 0}$ to ${\bf 3}$ are approximately 
$89.92\%$, $91.09\%$, $91.80\%$, and $89.93\%$, respectively. Comparing the 
results in Figs.~\ref{fig:2DCGL_D_th_LT}(b) and \ref{fig:2DCGL_D_th_LT}(c) 
with those in Figs.~\ref{fig:2DCGL_where}(b) and \ref{fig:2DCGL_where}(d), we 
observe a similar level of prediction accuracy, indicating that changing
the amplitude threshold for extreme events has little effect on predictability. 

Will modifying the labeling time duration parameter $p$ for extreme event 
have any effect on the predictability? To address this question, we double
the value of $p$ in Figs.~\ref{fig:2DCGL_when} and \ref{fig:2DCGL_where} 
from $10$ to $20$, so the prediction time horizon is now twice as long. 
In this case, there are $28524$ and $195486$ state images labeled as ${\bf 1}$ 
and ${\bf 0}$, respectively, where the number of states labeled as ${\bf 1}$ 
is larger due to the larger value of $p$. Figure~\ref{fig:2DCGL_D_th_LT}(d) 
shows the best ROC curves obtained from LeNet-5 and ResNet-50, where the 
respective AUC values are approximately $0.72$ and $0.82$. Comparing with 
the corresponding results in Figs.~\ref{fig:2DCGL_when}(b) and 
\ref{fig:2DCGL_when}(d), we see that a larger $p$ value tends to degrade 
the predictability of both types of DCNNs. This is intuitively expected: the 
time horizon of prediction can be increased but at the price of sacrificing 
accuracy. However, despite the decrease in accuracy, the AUC values for both 
cases are still much larger than $0.5$. Accuracy deterioration not only 
occurs in the time horizon of prediction, it is also reflected in the 
accuracy of prediction of the occurrence of the event in space.
Figure~\ref{fig:2DCGL_D_th_LT}(e) shows, for LeNet-5, the confusion matrix
for predicting ``where'' the extreme event will occur in one of the four 
subdomains, where the prediction accuracies for the four classes are 
approximately $53.8\%$, $55.8\%$, $58.2\%$, and $55.7\%$, respectively, which 
are lower than the corresponding values in Fig.~\ref{fig:2DCGL_where}(b). The 
results for ResNet-50 is shown Fig.~\ref{fig:2DCGL_D_th_LT}(f), where the 
prediction accuracies of the four classes are approximately $76.3\%$, 
$73.4\%$, $72.9\%$, and $75.5\%$, respectively, which are somewhat
lower than the corresponding accuracy values in Fig.~\ref{fig:2DCGL_where}(d). 
Overall, even with a doubly stretched prediction time horizon, both types of 
DCNNs are still capable of predicting ``when'' and ``where'' an extreme event 
will occur with reasonable accuracy, with the larger and deeper ResNet-50 
superior to LeNet-5 in terms of performance. [Detailed results (e.g., loss 
curves, ROC curves, AUC values, and confusion matrices) with a finer spatial 
grid ($5\times 5$ with 25 subdomains or classes) are presented in 
SI~\cite{SI}.] 

\begin{figure}[ht!]
\centering
\includegraphics[width=\linewidth]{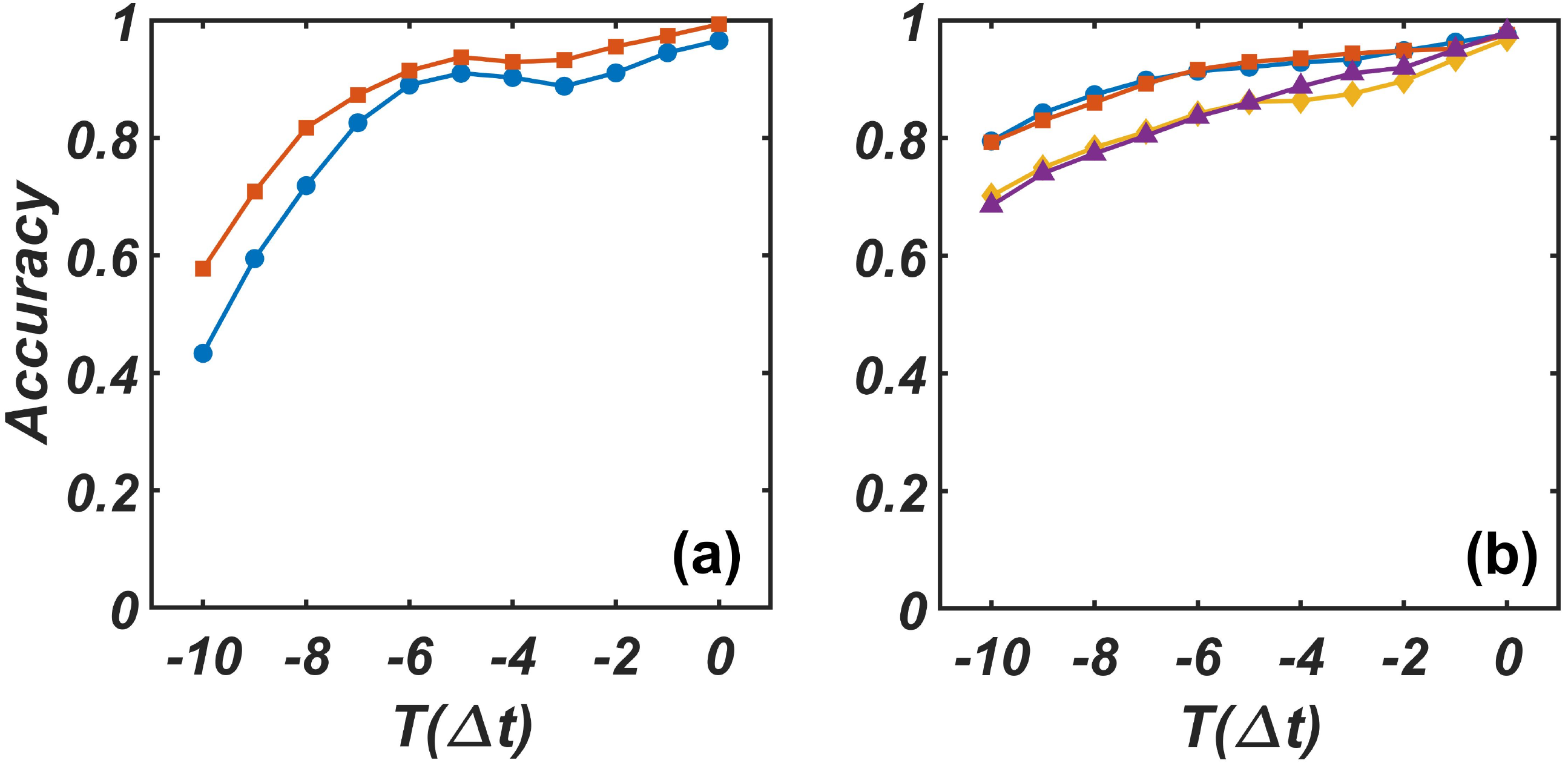}
\caption{ The ``when'' and ``where'' prediction accuracy at a 
series of time steps preceding the occurrence of an extreme event. In both
panels, the abscissa denotes the time step relative to the extreme event that 
occurs at time zero. For example, $-10$ means ten steps prior to the event. 
In (a) where the prediction scenario is ``when,'' the blue and brown curves 
represent the prediction accuracy at each time step for $|\bar{u}|=5.0$ and 
$|\bar{u}|=6.0$, respectively. The threshold used in calculating the ROC 
curve is 0.5. In (b), the prediction scenario is ``where,''  in which the 
blue, brown, yellow, and purple curves represent the cases of a $2\times 2$ 
spatial grid with $|\bar{u}| = 5$, a $2\times 2$ grid $|\bar{u}| = 6$, 
a $5\times 5$ grid with $|\bar{u}| = 5$, and a $5\times 5$ grid with 
$|\bar{u}| = 6$, respectively. The prediction horizon is $p=10$, and the
DCNNs are of the ResNet-50 type. The models for the results in panel (a) is 
the one in Figs.~\ref{fig:2DCGL_when}(d) and \ref{fig:2DCGL_D_th_LT}(a). 
For (b) the model is the one in Figs.~\ref{fig:2DCGL_where}(d), 
~\ref{fig:2DCGL_D_th_LT}(c), Fig.~S3, and Fig.~S5.}
\label{fig:2DCGL_ACC_dt}
\end{figure} 

\begin{figure}[ht!]
\centering
\includegraphics[width=\linewidth]{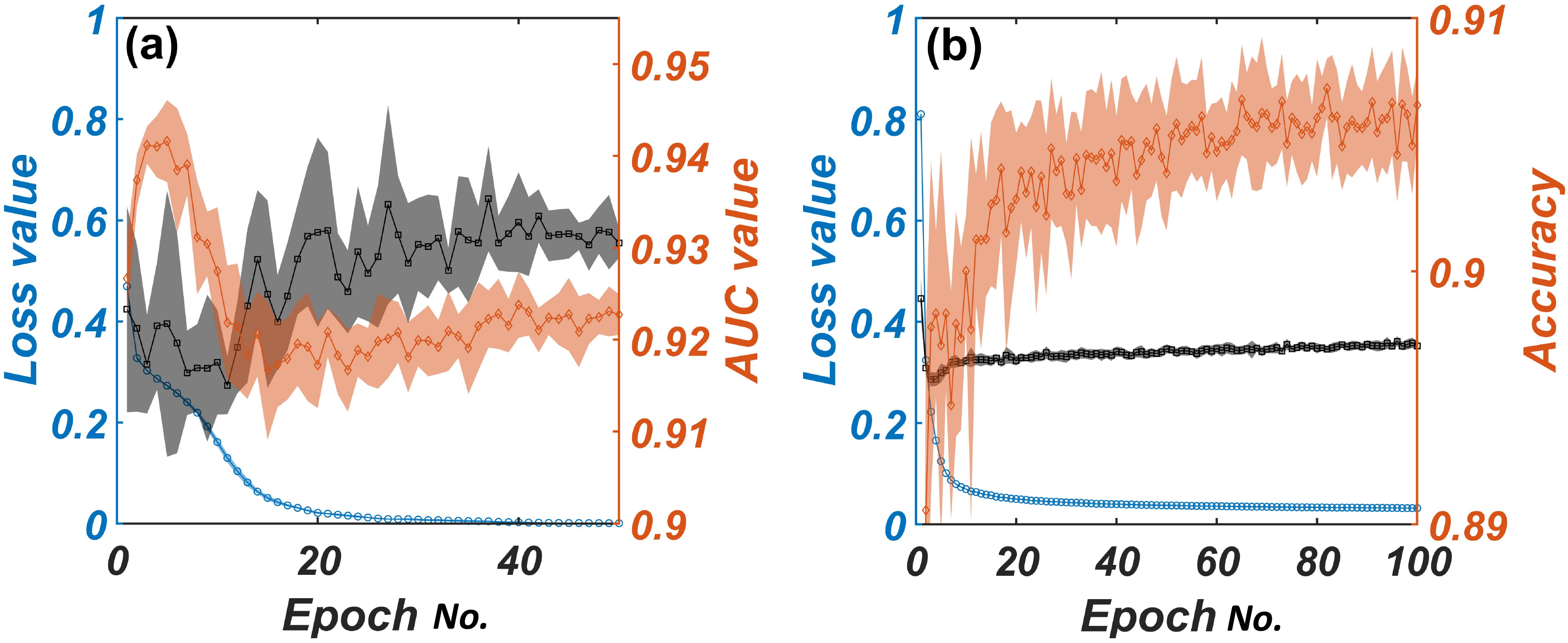}
\caption{Ensemble average of training loss, testing loss, AUC value, and accuracy for the ``when'' and ``where'' problems. (a,b) Results for the setting in Figs.~\ref{fig:2DCGL_when}(c) and \ref{fig:2DCGL_where}(c), respectively. In (a), the blue, black, and brown traces correspond to the ensemble average of training loss, testing loss, and AUC value, respectively. The shadowed region about each line is within the standard deviation. In (b), the blue, black, and brown traces correspond to the ensemble average of training loss, testing loss, and accuracy, respectively.}
\label{fig:Ensemble_CGLE}
\end{figure}

\begin{figure*}[ht!]
\centering
\includegraphics[width=\linewidth]{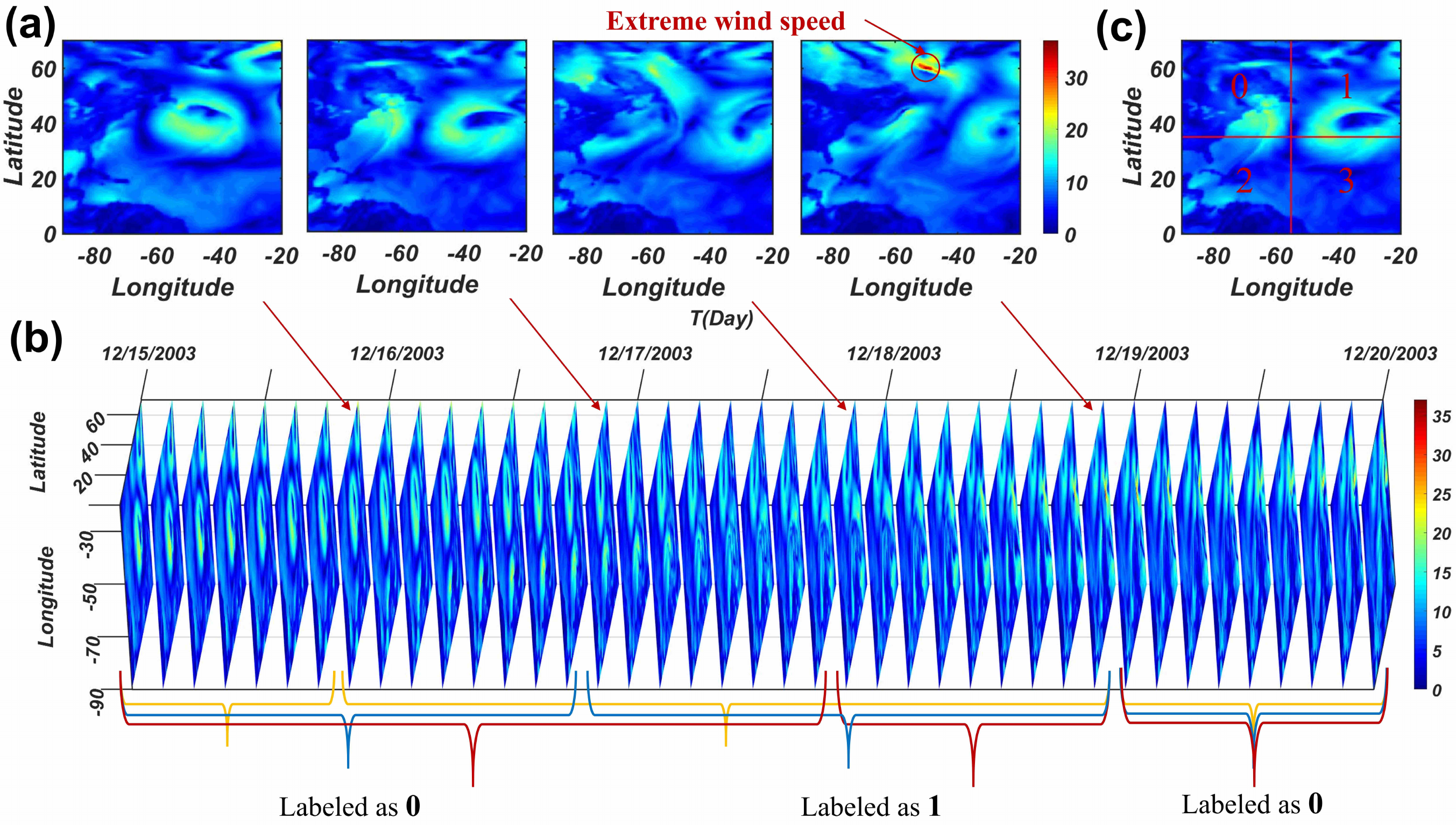}
\caption{Image data of wind speed in the North Atlantic ocean and labeling 
schemes for DCNN. (a) Four representative images of the wind speed in the 
North Atlantic ocean and (b) a stack of wind speed images from December 15 
to 20, 2003. In (a), the horizontal and vertical axes are the west 
longitude and north latitude, respectively, and the time instants that the 
four states occur are marked in (b). Three labeling schemes for using DCNNs
to predict ``when'' an event of extreme wind speed will occur are specified 
in (b). (c) A labeling scheme for predicting ``where'' an extreme event will 
occur.}
\label{fig:EWE_LS_P}
\end{figure*}

To assess the performance of DCNN within the horizon of predicting ``when''
and ``where'' an extreme event would occur, we calculate the accuracy at each
time step within the prediction horizon, as shown in
Figs.~\ref{fig:2DCGL_ACC_dt}(a) and ~\ref{fig:2DCGL_ACC_dt}(b), respectively. 
In particular, Fig.~\ref{fig:2DCGL_ACC_dt}(a) shows that, ten time steps prior 
to the occurrence of the extreme event, the prediction accuracy is the lowest. 
This is expected because the earlier to start prediction, less is the 
information about the extreme event. At this time, for $|\bar{u}| = 6$, the 
prediction accuracy is about $60\%$, which is already above the $50\%$ random 
level. Moving closer to the event, the prediction accuracy tends to increase 
toward $100\%$. For example, at seven time steps before the extreme event, the
accuracy exceeds $80\%$ for both $|\bar{u}| = 5$ and $|\bar{u}| = 6$ cases.
Figure~\ref{fig:2DCGL_ACC_dt}(b) shows the accuracy of predicting ``where'' an 
extreme event will occur at each time step within the prediction horizon. For 
the two cases of partitioning the domain into $4$ subregions, the prediction 
accuracy is at or above the $80\%$ level at each time step. For the two cases 
of $25$ subregions, the overall prediction accuracy is reduced slightly but it 
is still at or above the $70\%$ level for each time step. 
Figure~\ref{fig:2DCGL_ACC_dt} thus demonstrates that DCNNs are capable of 
predicting ``when'' and ``where'' an extreme event will occur within the 
prediction horizon.

To test the stability of our DCNN prediction method, we take the 2D CGLE system and produce ten random realizations of the training and test using the same settings as in Figs.~\ref{fig:2DCGL_when}(c) and \ref{fig:2DCGL_where}(c). Figures~\ref{fig:Ensemble_CGLE}(a) and \ref{fig:Ensemble_CGLE}(b) show the ensemble average of training loss, testing loss, the AUC value, and the prediction accuracy for the settings of Figs.~\ref{fig:2DCGL_when}(c) and \ref{fig:2DCGL_where}(c), respectively. It can be seen that all ten realizations of the training loss, testing loss, AUC value, and accuracy exhibit the same trend. For example, when using ResNet-50 to predict ``when'' an extreme event would occur, the largest AUC value is achieved at about five epochs, as shown in Fig.~\ref{fig:Ensemble_CGLE}(a). When predicting the location of the extreme event, the prediction accuracy approaches a constant value as more training epochs are used, as shown in Fig.~\ref{fig:Ensemble_CGLE}(b). The largest average prediction accuracy occurs at about the $82$nd epoch. Therefore, within 100 training epochs, there is little over-fitting, demonstrating that the DCNN is able to generate stable and accurate prediction results.
    
\subsection{Predicting the occurrence of extreme wind speed in North Atlantic 
ocean based on real-world data}

We test the power of DCNNs to predict extreme events in the real world.
Our concrete example is to predict the extreme wind speed in the North 
Atlantic ocean from real datasets (described in Appendix~\ref{appendix_C}). 
Figure~\ref{fig:EWE_LS_P}(a) shows four typical wind speed images, where an 
extreme event of wind speed greater than $30$m/s occurs towards the northern 
part in the rightmost image as indicated and the three images from left to 
right represent the data three days, two days, and one day before the 
occurrence of the extreme event, respectively. The occurrence time of the 
four data images is indicated in Fig.~\ref{fig:EWE_LS_P}(b). Three possible 
labeling schemes for a DCNN to predict ``when'' an extreme event will occur 
are specified in Fig.~\ref{fig:EWE_LS_P}(b) by the red, blue and yellow 
brackets, in which the data one, two, and three days prior to the occurrence 
of the extreme event are labeled as ${\bf 1}$, corresponding to a prediction 
horizon of one, two, and three days, respectively. For convenience, we call 
the three corresponding labeling schemes as 3DLS, 2DLS, and 1DLS. In general, 
setting a longer horizon leads to reduced prediction accuracy. 
(The label shown in Fig.~\ref{fig:EWE_LS_P}(b) is only related to the extreme 
wind event in ~\ref{fig:EWE_LS_P}(a). The actual label scheme takes into 
account the situation where two extreme events occur in the same prediction
horizon. The time evolution of the maximum wind speed is shown in Fig.~S1 in
Supporting Information.
Figure~\ref{fig:EWE_LS_P}(c) depicts a labeling scheme for predicting 
``where'' the extreme event will occur, after it has been predicted to occur 
within a specific time horizon, where the spatial domain of interest is 
divided into $4$ sub-domains with distinctive labels, leading to $4$ 
classes to be trained and predicted. 

\begin{figure}[ht!]
\centering
\includegraphics[width=\linewidth]{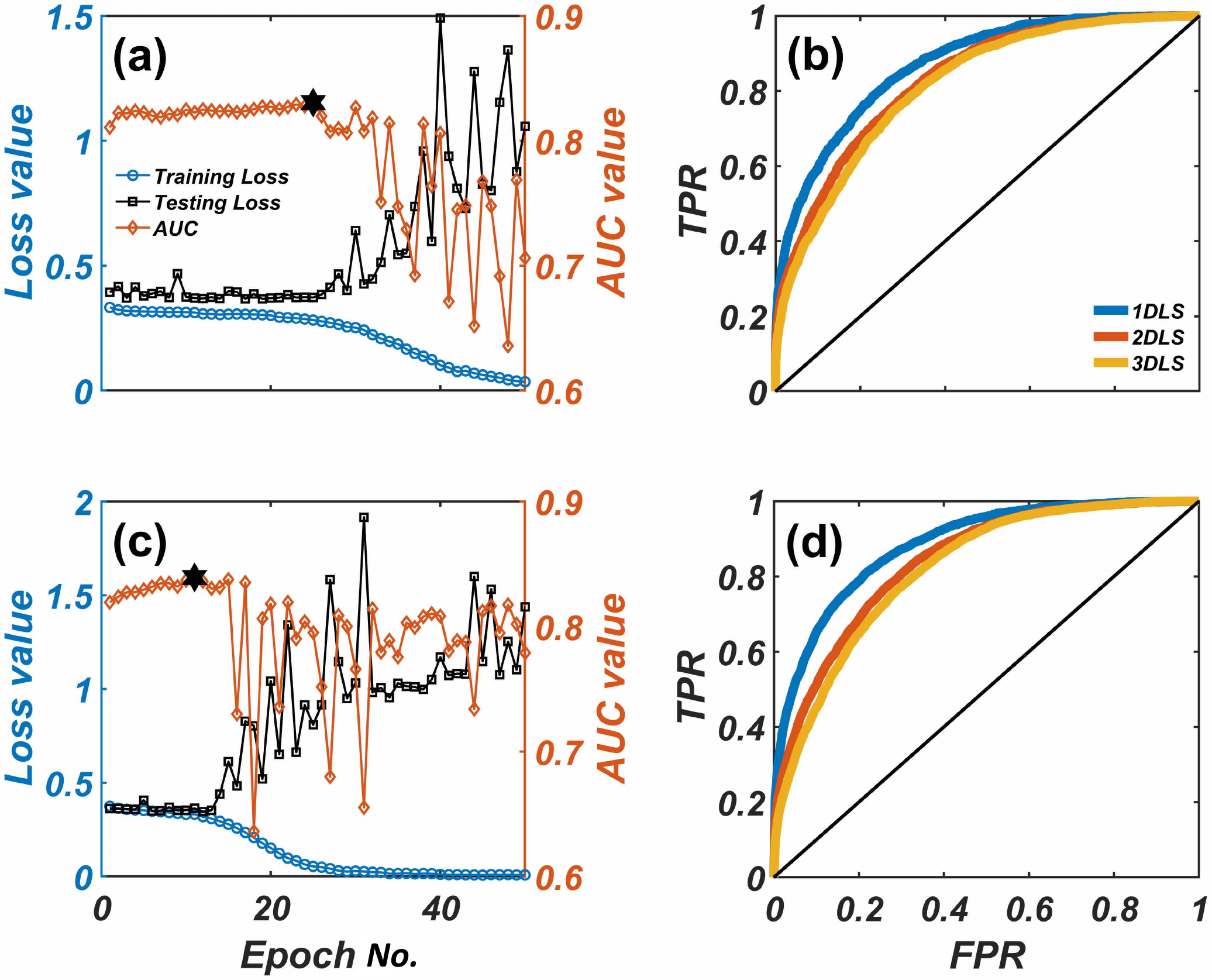}
\caption{DCNN based prediction of ``when'' an extreme wind speed will occur. 
(a,b) Losses, AUC value, and ROC curves for the dataset restructured from 
the original data with time step $\Delta t = 6$ hours. (c,d) The same quantities
as those in (a,b) but for the forecast dataset with time step $\Delta t = 3$ 
hours. In (a,c), training and testing losses as well as changes in the AUC 
value are plotted with the increase of training epochs for 2DLS. (b,d) The 
ROC curves for the three distinct labeling schemes specified in 
Fig.~\ref{fig:EWE_LS_P}(b). In all cases, the DCNN is ResNet-50 and the 
threshold value of the wind speed for defining an extreme event is set as 
$30$ m/s.}
\label{fig:EWE_when}
\end{figure}

\begin{table}[h!]
\begin{center}
\caption{Table of AUC values for different datasets and labeling schemes.}
\label{tab:table_AUC}
\begin{tabular}{|l|l|l|l|} 
\hline
~ & 1DLS & 2DLS & 3DLS \\
\hline
Restructured dataset & 0.869 & 0.831& 0.816\\
\hline
Forecast dataset & 0.884 & 0.839& 0.820\\
\hline
\end{tabular}
\end{center}
\end{table}

Employing the labeling scheme in Fig.~\ref{fig:EWE_LS_P}(b), we present 
results with ResNet-50 for its ability to predict ``when'' an extreme event 
will occur (our results with the CGLE indicate that ResNet-50 generally has 
better performance than LeNet-5), as shown in Fig.~\ref{fig:EWE_when}.We use 
the first $30$ and the last $10$ years of wind speed data as the training 
and testing datasets, respectively, and the total numbers of images in the 
training and testing datasets are $43830$ and $14610$ for the restructured
data, and $87660$ and $29220$ for the forecast data (Appendix~\ref{appendix_C}).
Figures~\ref{fig:EWE_when}(a) and \ref{fig:EWE_when}(c) show that the 
training and testing losses decrease simultaneously in the first $25$ and 
$11$ epochs before over-fitting occurs. 
For the forecast dataset with 2DLS, the best performance 
is achieved for $\mbox{epoch}=11$. However, for the restructured dataset with 
2DLS, the optimal time occurs at $\mbox{epoch}=25$. The best AUC values
for the restructured and forecast datasets are approximately $0.83$ and 
$0.84$, respectively. Since the only difference between the two kinds of 
datasets is the sampling time, the similar AUC values obtained means that 
the time resolution of the dataset has little effect on the prediction 
accuracy. Figure~\ref{fig:EWE_when}(b) shows, for the restructured dataset, 
the ROC curves from the three labeling schemes, where the AUC value for 
1DLS, 2DLS, and 3DLS are approximately $0.869$, $0.831$, and $0.816$, 
respectively. As expected, as the prediction horizon expands, the accuracy
decreases. However, remarkably, even when the horizon becomes three times 
as long, the decrease in the accuracy is quite insignificant. Similar results
have been obtained with the forecast dataset, where the AUC value for 1DLS, 
2DLS, and 3DLS are approximately $0.884$, $0.839$, and $0.820$, respectively,
as shown in Fig.~\ref{fig:EWE_when}(d). 
The AUC values are summarized in Tab.~\ref{tab:table_AUC}. 
These results indicate the strong ability and reliability of the DCNN for 
predicting ``when'' an extreme wind speed will occur even three days in 
advance. 

\begin{figure}[ht!]
\centering
\includegraphics[width=\linewidth]{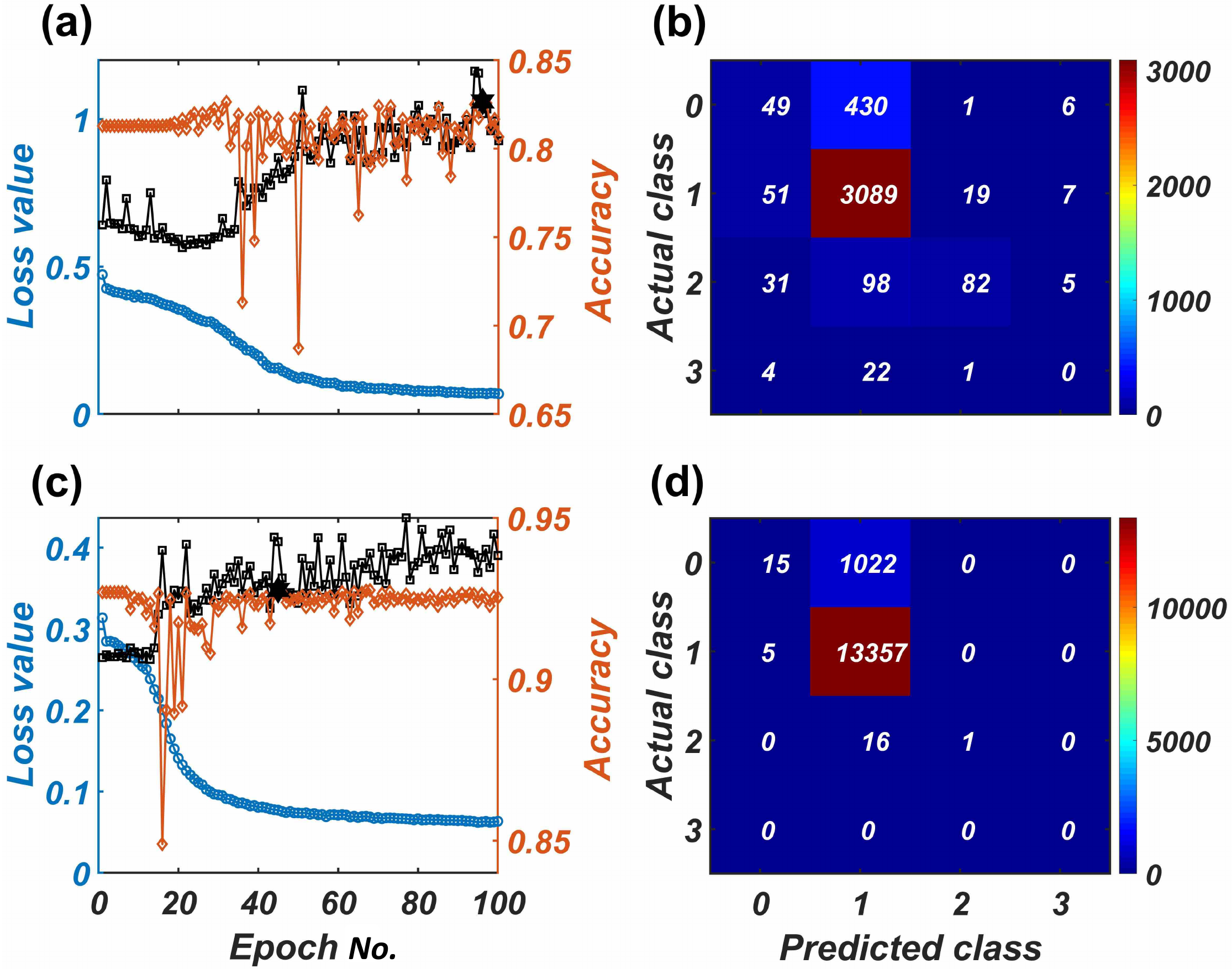}
\caption{Results of DCNN based prediction of ``where'' an extreme wind speed 
will occur. (a) Losses (blue and black) and accuracy (brown) versus the epoch 
number for the restructured dataset, and (b) the confusion matrix corresponding 
to the best accuracy. (c,d) Results similar to those in (a,b) but for the 
forecast dataset. In all cases, the DCNN is ResNet-50, the labeling scheme is 
2DLS, and the threshold in the wind speed for defining an extreme event is 
$30$m/s.}
\label{fig:EWE_where}
\end{figure}

With the labeling scheme in Fig.~\ref{fig:EWE_LS_P}(c), we can predict the
location of the occurrence of an extreme wind speed after it has been 
predicted to occur within the prediction horizon. Representative results 
are shown in Fig.~\ref{fig:EWE_where}. 
As shown in Fig.~\ref{fig:EWE_LS_P}, an image contains an 
extreme event if the 10-meter wind speed exceeds $30$ m/s.
For two or more successive occurrences of extreme wind speed, we pick out 
the images corresponding to each event. This can cause duplication of some 
images in the training and testing datasets. The whole restructured and 
forecast datasets have $15579$ and $57664$ images, respectively, where the 
large difference is due to the occurrences of many consecutive extreme events. 
The restructured dataset has $1458$, $13482$, $594$, and $45$ wind speed 
images for classes ${\bf 0}$, ${\bf 1}$, ${\bf 2}$, and ${\bf 3}$, 
respectively, while the forecast 
datasets has $4114$, $53346$, $170$, and $34$ images for classes ${\bf 0}$, 
${\bf 1}$, ${\bf 2}$, and ${\bf 3}$, respectively. The wind speed datasets 
are inevitably biased because there is an unbalanced distribution of extreme 
events in the North Atlantic ocean. For both the restructured and forecast 
datasets, we use the first $3/4$ as the training dataset and the remaining 
$1/4$ as the testing set. Figure~\ref{fig:EWE_where}(a) presents results
of losses and accuracy for the restructured dataset, with the confusion matrix
corresponding to the best accuracy is shown in Fig.~\ref{fig:EWE_where}(b).
We see that the DCNN predictions are quite successful for classes $0$ and $2$, 
although most of the predicted events are for class $1$. The overall accuracy 
is about $82.7\%$. Figures~\ref{fig:EWE_where}(c) and \ref{fig:EWE_where}(d) 
show the corresponding results but for the forecast dataset, where the overall 
accuracy is approximately $92.8\%$. The marked increase in the accuracy is
due to the fact that more extreme events in the forecast dataset belong to 
class $1$ than those in the restructured dataset. The DCNN also predicts 
$15$ and $1$ correct images for classes ${\bf 0}$ and ${\bf 2}$, respectively. 
Taken together, our tests indicate that DCNN can be quite effective at 
predicting not only ``when'' extreme wind speed will occur within a predefined 
time interval but also ``where'' it will occur with reasonable accuracies. 
(In SI~\cite{SI}, we present the confusion matrices for a labeling scheme 
based on a $5\times 5$ spatial grid with $25$ classes, the ``where'' results 
for restructured and forecast datasets with no data overlap in the training 
and testing data, and the results of predicting ``when'' and ``where'' with 
the critical wind speed for defining an extreme event reduced to 
$26$ m/s.)

\begin{figure}[ht!]
\centering
\includegraphics[width=\linewidth]{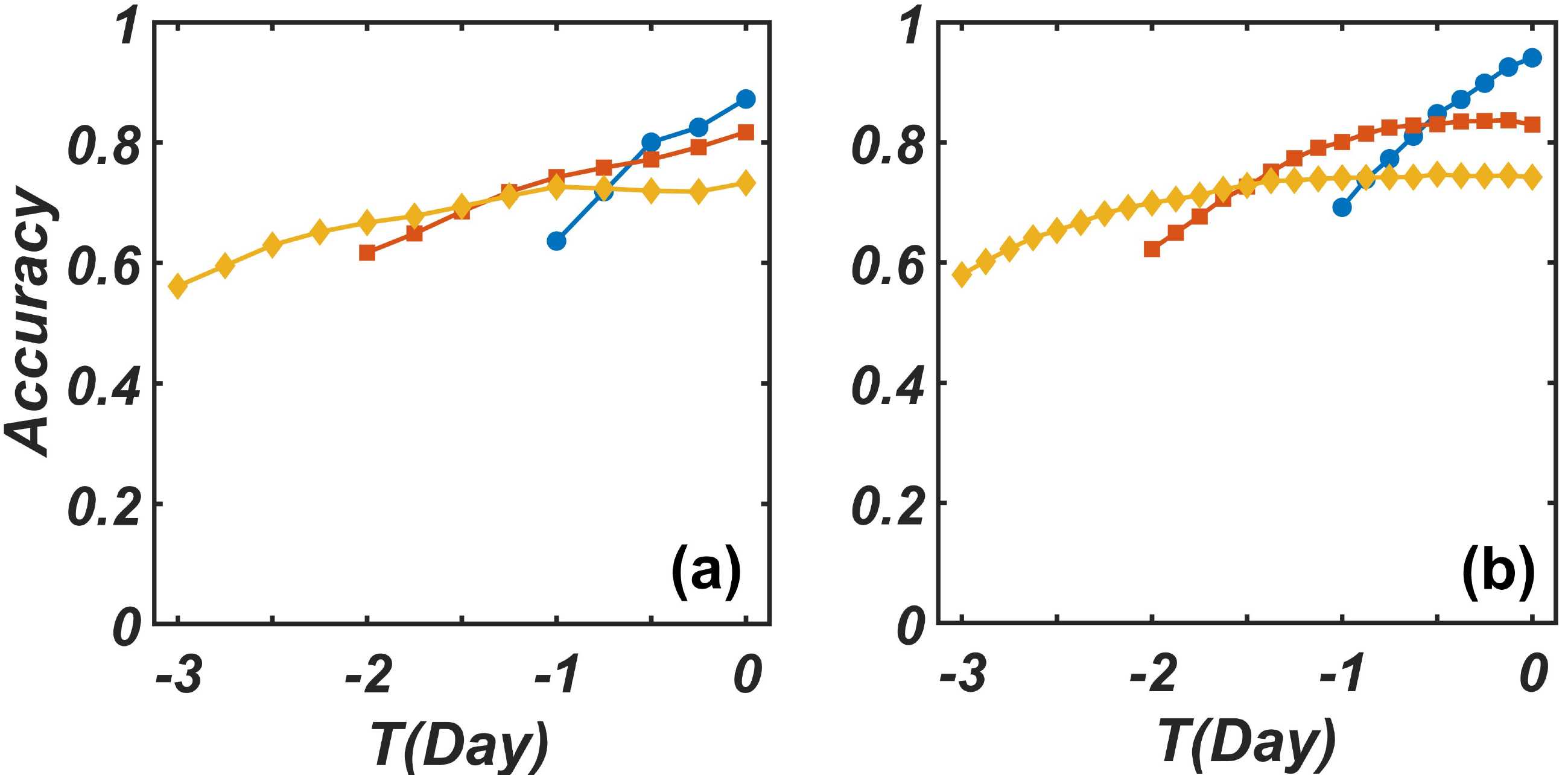}
\caption{ 
Accuracy of predicting when an extreme wind speed will occur at a series of time steps preceding the occurrence. In both panels, the abscissa denotes the time relative to the extreme event that occurs at time zero. For example, $-3$ means three days prior to the event. (a) For the restructured dataset, the prediction accuracy at each time step for 1DLS, 2DLS, and 3DLS (represented by the blue, brown, and yellow curves, respectively). (b) The same legends as in (a) but for the forecast dataset. The curves are generated with the threshold $FPR=0.15$ on the ROC curve. The wind speed threshold for extreme wind events is $30m/s$ and the DCNNs are of the ResNet-50 type. The parameters for the results in panels (a) and (b) are the same as those in Fig.~\ref{fig:EWE_when}(b) and Fig.~\ref{fig:EWE_when}(d), respectively.}
\label{fig:WSD_ACC_dt}
\end{figure} 

To evaluate the performance of DCNN within the horizon of predicting ``when'' an extreme wind speed event would occur, we calculate the accuracy at each time step within the prediction horizon for both the restructured and forecast datasets, as shown in Figs.~\ref{fig:WSD_ACC_dt}(a) and ~\ref{fig:WSD_ACC_dt}(b), respectively. It can be seen that the 1DLS labeling scheme has the best predictability for the state close to the extreme wind event, but the prediction accuracy decays quickly with the number of time steps prior to the extreme event. On the contrary, the labeling scheme 3DLS has the slowest decay in the prediction accuracy, because training the DCNN with more states can prolong the prediction horizon (in this case, to three days before the extreme event). However, the inclusion of more wind speed states decreases the importance of features of the states that are close to the extreme event, resulting in a low prediction accuracy for both the reconstructed and forecast datasets. Nevertheless, the prediction accuracy is still larger than the chance level 0.5. Figure~\ref{fig:WSD_ACC_dt} also reveals that the prediction accuracy is not one at time $T=0$, demonstrating that the DCNN does not simply rely on a threshold to predict the extreme wind speed event. Overall, the results in Fig.~\ref{fig:WSD_ACC_dt} indicate that the DCNN is capable of predicting ``when'' an extreme wind speed event will occur three days in advance reasonably accurately.

\section{Discussion}

We have formulated the challenging problem of model-free prediction of 
extreme events in physical situations where a nonlinear dynamical process 
occurs in a two-dimensional spatial domain as an image 
recognition/classification problem in machine learning. While the 
spatiotemporal evolution of the physical field in such a 
domain is governed by 2D nonlinear PDEs, our framework requires no model and 
is completely data based. We have exploited DCNNs and demonstrated, using 
synthetic data from the CGLE and real empirical data of wind speed in the 
North Atlantic ocean, that when and where in the spatial domain extreme 
events would occur can be predicted with reasonable accuracies. Especially,
we break the problem of spatiotemporal prediction into two parts: when 
and where an extreme event will occur. To address the issue of ``when'',
we set a desired prediction horizon in time, based on which spatiotemporal
data in the form of consecutive snapshots (images) can be properly labeled. 
A DCNN successfully trained with a large number of labeled snapshots is then
capable of predicting the occurrence of the extreme event within the 
predefined time horizon. Intuitively, there is a trade-off between the 
prediction horizon and accuracy: a longer horizon will inevitably reduce the 
accuracy, which has been demonstrated. The ``where'' problem can be solved
after it has been predicted that an extreme event will occur within the 
prediction horizon, again through a proper labeling scheme based on a uniform
division of the spatial domain. Another trade-off arises: the prediction 
accuracy will decrease with an increase in the resolution of the spatial 
domain leading to more classes of images to be predicted. We have tested
two types of DCNN: LeNet-5 and ResNet-50, and found that the latter, a deeper 
neural networked system, performs relatively better.

We have studied two types of DCNN in this paper: LeNet-5 and ResNet-50, with 
the general problem setting of predicting to which class a state image belongs. 
This method can also be adopted to addressing image segmentation problems, 
where it is necessary to change the class in our labeling scheme to a 
representation mask. For example, class $0$ in predicting ``where'' an extreme 
event will occur by partitioning the spatial domain of interest to $4$ 
subdomains can be represented as a mask as 
$\begin{vmatrix}
1 & 0 \\
0 & 0 
\end{vmatrix}$. 
DCNNs have been used widely in image segmentation, e.g., fully convolutional 
networks~\cite{LSD:2015} and U-Net~\cite{RFB:2015}. To exploit the DCNNs for 
image segmentation to predict ``when'' and ``where'' extreme events occur is 
an interesting topic of research.

For predicting extreme events in the CGLE, the accuracy achieved is quite
remarkable and is generally markedly higher than predicting the extreme wind 
speed from empirical data, especially when predicting the location of its 
occurrence. The main culprit for the reduced accuracy with the real data lies
in the inhomogeneous or biased occurrence of the extreme event in space, i.e., 
the probability of its occurrence is not uniform in the domain, in contrast to 
systems described by the CGLE where extreme events occur approximately 
uniformly in space. To develop a machine learning framework to predict 
extreme events in a 2D spatial domain with biased data is an issue that must 
be addressed in real applications, calling for further efforts in this field.

When applying the DCNN based prediction scheme to the wind speed datasets,
in addition to the spatially inhomogeneous or biased occurrences of extreme
wind speed event, the lack of sufficient extreme wind speed events is also a 
problem, as both can lead to the fast over-fitting of the DCNNs. With more 
available data together with fine-tuning the hyperparameters, the overfitting 
problem can be mitigated or even eliminated to enhance prediction performance.

The DCNN framework developed in this paper is anticipated to be applicable to
real-world systems, for two reasons. First, DCNNs have enjoyed dramatic,
unprecedented successes in many fields of science and engineering for complex
tasks ranging from image recognition and classification to facial recognition 
to EI Ni\~no/South Oscillation prediction. The basic idea underlying our 
prediction framework is that extreme events in 2D spatial domains can naturally
be represented as ``hot spots'' in images, rendering the DCNN approach 
applicable. Second, DCNNs themselves are complex, high-dimensional nonlinear 
systems. Intuitively, insofar as properly DCNNs are able to overpower the      
target spatiotemporal dynamical systems in complexity, in principle the former
will have the predictive power over the latter. In spite of the remarkable
successes of DCNNs in solving all kinds of previously unsolvable problems, an
open challenge is to understand its inner workings, giving birth to the field
``Explainable Machine Learning.'' At the present, the underlying mechanisms of
DCNNs have not been understood, but this lack of understanding has not hindered 
the exploitation of DCNNs for solving sophisticated problems. In fact,     
designing suitable DCNNs to solve challenging problems, while postponing the
formidable issue of gaining some basic understanding, has been a widely
accepted practice in the scientific community. In addition to the type of
spatiotemporal chaotic systems treated in this paper, there are other types of
complex and chaotic systems exhibiting extreme events. Insofar as the available
data from these systems can be organized as 2D images, our DCNN approach should
generally be applicable. In particular, for any specific nonlinear system, the 
DCNNs need to be trained to learn the intrinsic rules governing the dynamical
evolution of the images to make predictions.

It can happen that two extreme events occur within the same prediction horizon.
We find that, when choosing the horizon to be $10$ time steps, such cases are 
rare (e.g., less than $4\%$ in the 2D CGLE system). If such a situation does
arise, it will cause a small reduction in the prediction accuracy, due to the
necessity to assign labels to images prior to the occurrence of the extreme
events. In particular, to predict ``when'' an extreme event would occur, we 
label the images within the prediction horizon of an extreme event as one. If 
two extreme events occur during the same horizon, the images between these two
extreme events are still labeled as one. However, to predict the spatial
location of the extreme event (the ``where'' problem), the situation is
slightly more complicated. Suppose two extreme events occurring within the
same horizon belong to the same subdomain $\chi$. In this case, we duplicate
the overlapping images between the two extreme events and directly label them
and their replica as $\chi$. If two extreme events occurring in the same
horizon belong to two different subdomains $\chi_1$ and $\chi_2$, 
we duplicate the overlapping images between these two extreme events and label
them and their replica as $\chi_1$ and $\chi_2$. Inevitably, this labeling
scheme will lead to errors as we assign some identical images to two distinct
classes. However, the fraction of the overlapping images in the total dataset
is negligibly small, leading to little reduction in the prediction accuracy.

In employing DCNNs to predict ``when'' and ``where'' extreme events occur 
in the CGLE with the demonstrated success, the temporal information about the 
dynamical evolution of the 2D state has in fact not been adequately exploited,
as only static 2D images at a discrete set of time are used for training and
prediction. For an input image, the DCNN machine only gives a time interval, 
with the present time as the starting point, during which an extreme event 
occurs. To develop a machine learning framework to take full advantage of 
the temporal information to achieve more reliable and accurate prediction is 
an outstanding problem.
 
Recently, there have been advances in exploiting reservoir computing machines, 
a class of recurrent neural networks articulated about two decades 
ago~\cite{Jaeger:2001,JH:2004}, to predict the state evolution in 
spatiotemporal dynamical systems described by nonlinear 
PDEs~\cite{HSRFG:2015,LBMUCJ:2017,PHGLO:2018,Carroll:2018,NS:2018,ZP:2018,
WYGZS:2019,JL:2019,VPHSGOK:2020,FJZWL:2020,KFGL:2021a} for about half dozen 
Lyapunov time. It thus seems possible to develop a framework based on reservoir
computing to predict the precise time of occurrence of extreme events. We have 
conducted a pilot study of this idea with the CGLE, but the results obtained 
so far have not been promising. A plausible reason is that extreme events are 
relatively rare: starting from a random initial condition, it usually requires 
the system to evolve for a time much longer than half dozen Lyapunov times for 
an extreme event to occur. In fact, in spite of the deterministic nature of the
CGLE system, the extreme events appear to occur randomly in both space and time.
Since reservoir computing only involves one hidden layer and is designed to 
train a neural network to predict the state evolution of the target system, it 
may not be effective to deal with extreme events that are ``quasi''-stochastic.
In contrast, DCNNs, because of their sophisticated structure and vast 
complexity, are capable of predicting extreme events.

Another relevant issue concerns about the time scale of the target system 
and the determination of the prediction horizon. 
The parameter region in which the 2D CGLE system exhibits extreme events can 
have chaos in both space and time. The mean temporal period is an important
characteristic time scale of the system. However, our DCNN based prediction
framework should be applicable even to systems without a clearly defined time
scale, because DCNNs are specially designed to recognize random spatial 
patterns. If the generation of an extreme event leads to the buildup of some 
distinct spatial pattern, then DCNN can be effective, regardless of whether the
underlying dynamical system has a mean temporal period. The determination 
of the prediction horizon is also crucial. For systems
with a natural cycle, it can be chosen as the prediction horizon. If a system
does not have such a natural cycle, then the average correlation time or 
another physically meaningful time can be used. For example, for the North 
Atlantic wind speed dataset, the relevant time is one day or a few days. It 
is indeed desired to develop a more rigorous and theoretically justified 
criterion to determine the prediction horizon.

\section*{Acknowledgments}

The work at Arizona State University was supported by AFOSR under Grant 
No.~FA9550-21-1-0438 and by ONR under Grant No.~N00014-21-1-2323. The work
at Xi’an Jiaotong University was supported by the National Key R $\&$ D 
Program of China (2021ZD0201300), National Natural Science Foundation of 
China (No.~11975178), and K.~C.~Wong Education Foundation. 

\appendix

\section{Two-dimensional complex Ginzburg-Landau equation} \label{appendix_A}

Mathematically, the 2D CGLE is given by
\begin{align} \label{eq:CGLE}
\cfrac{\partial u}{\partial t} = \epsilon u + (\gamma + i\alpha)|u|^2u - (\mu + i\beta) \nabla^2 u,
\end{align}
where $\epsilon$, $\gamma$, $\mu$, $\alpha$ and $\beta$ are dimensionless
parameters. For $\epsilon = 0$, $\gamma = 0$, and $\mu = 0$,
the CGLE is reduced to the nonlinear Schr\"{o}dinger equation (NLSE):
\begin{align} \label{eq:NSE}
\cfrac{\partial u}{\partial t} = + i\alpha |u|^2 u - i\beta \nabla^2 u,
\end{align}
whose solutions diverge for $\alpha\beta > 0$. In particular, for
$\alpha\beta > 1$, all solutions of the NLSE are unstable. In the
parameter regime $(\gamma,\epsilon,\mu) \ll (\alpha,\beta)$, the regime
in which the CGLE is close to the nonlinear Schr\"{o}dinger equation, the
system exhibits an intermittently bursting behavior in both space and time.
From the solutions of the CGLE in the parameter regime near the NLSE limit,
extreme or rare intense events in the form of spikes or bursts with
``unusually" large amplitude can occur. The CGLE in this parameter regime
thus represents a mathematical paradigm for studying extreme events in
spatiotemporal dynamical systems. 

In our study, we set $\epsilon=1$, $\gamma=1$, $\alpha=-30$, and $\beta=30$, 
so that the system generates extreme events that are random in both space 
and time. We solve the spatiotemporal evolution of the 2D state $u(x,y,t)$ on 
a square region of side length $l=10\pi$ in the $(x,y)$ plane with periodic 
boundary conditions using the pseudospectral and exponential-time differencing 
scheme~\cite{CM:2002}. Spatial discretization is accomplished by covering the 
domain with a uniform, $100\times 100$ grid. The integration time step is 
$dt=10^{-4}$. We perform sampling in the time domain on the numerical solution 
with $\Delta t=0.01$ to obtain a manageable set of 2D state images. For the 
parameter setting, the average temporal and spatial periods are approximately 
0.09 and 5.09, respectively, corresponding to about $9$ time steps and the
size of $16$ spatial cells in the resampled data set. 
Besides, the largest Lyapunov exponent of the system in this parameter setting
is estimated to be $\lambda_{max}\approx3.3$, so the Lyapunov time is
$\Lambda_t=1/(\lambda_{max})\approx0.3$. For the choice $\Delta t = 0.01$,
the prediction horizon is $10\Delta t = 0.1$, which is within one Lyapunov
time, rendering predictable the extreme event.

We also calculate the temporal correlation function $C_{st}(\tau)$ for the 2D 
CGLE system, which is defined as
\begin{equation}
C_{st}(\tau)=\langle\frac{\langle|u(x,y,t)u^*(x,y,t+\tau)|\rangle_t}{\langle|u(x,y,t)|^2\rangle_t}\rangle_s,
\end{equation}
where $^*$ denotes the complex conjugate, $\langle\cdot\rangle_t$ and
$\langle\cdot\rangle_s$ are averages over time and space, respectively. We 
use $10^4$ sampling time steps to calculate the average over time. The result 
is shown in Fig.~\ref{fig:2DCGL_P}(f), where the decay time constant is about 
10 time steps.

In our study, it took about nine hours for training $100$ epochs on ResNet-50 to predict when an extreme event would occur in the 2D CGLE on an NVIDIA GeForce RTX 3080 Laptop GPU. For the ``where'' problem, under the same setting, the time required for training was about two hours.

\section{Calculation of ROC curve} \label{appendix_B}

The detection of extreme events can be viewed as a binary classification 
problem. Let $x$ be the maximum field intensity in space, e.g., $|u|_{max}$ in 
the 2D CGLE or the maximum wind speed in North Atlantic ocean. For actual
extreme events, $x$ follows a probability distribution with the density 
function $f_1(x)$. When, at a specific time instant, there is no extreme
event in the space, then $x$ follows another probability distribution with 
the density function $f_0(x)$. By definition, on the $x$-axis, the center of
$f_1(x)$ must be on the right side of the center of $f_0(x)$. To classify if 
a snapshot image possesses an extreme event, we set up a threshold
value $X$, where if $x > X$ ($x < X$), then $x$ is regarded as belonging 
to $f_1(x)$ [$f_0(x)$]. With respect to a specific threshold value $X$, the 
true positive rate and false positive rate are given by 
\begin{align} \nonumber
\mbox{TPR}(X) &= \int^{\infty}_{X} f_1(x) dx \ \ \mbox{and} \\ \nonumber
\mbox{FPR}(X) &= \int^{\infty}_{X} f_0(x) dx,
\end{align}
respectively. Let $[X_1,X_2]$ be the range outside of which both $f_0(x)$ and
$f_1(x)$ are zero. As $X$ is decreased from $X_2$ to $X_1$, both TPR and FPR 
continuously increase from zero to one.  

In an actual situation, the probability density functions $f_0(x)$ and $f_1(x)$
are not known, so the TPR and FPR must be evaluated approximately by the 
relative frequencies. From our testing data consisting of a large number of
images, the ground truth, i.e., whether an image contains an extreme event, is 
known. For a fixed threshold value $X$, the prediction of an extreme event from
an image is either true or a false alarm. By sifting through all the images
in the testing dataset, we can estimate the TPR and FPR values. Varying 
$X$ continuously in a proper range leads to systematically changing values 
of TPR and FPR, thereby generating the ROC curve. In our work, we use the 
MatLab function ``perfcurve'' to calculate the various ROC curves. 

\section{Wind-speed datasets of North Atlantic ocean} \label{appendix_C}

We use the ERA-Interim datasets from European Centre for Medium-Range Weather 
Forecasts (ECMWF), downloaded from 
\href{https://www.ecmwf.int/en/forecasts/datasets/reanalysis-datasets/era-interim}{https://www.ecmwf.int}. The details of the ERA-Interim datasets can be 
found in Ref.~\cite{DUSBPKABBB:2011}. From the datasets, we extract 
10-meter wind speed (wind speed at a height of 10 meters) data of the North 
Atlantic ocean, with $\mbox{latitude} \in [0^{\circ}N, 70^{\circ}N]$ and 
$\mbox{longitude} \in [20^{\circ}W, 90^{\circ}W]$. The dataset has a 40-year
duration: from January 1, 1979 to December 31, 2018. There are two types of 
datasets: the restructured dataset with time interval of six hours and $58440$ 
wind-speed images in total, and the forecast dataset with time interval of
three hours and $116880$ images. For both datasets, we set two wind-speed 
thresholds to define extreme events: $26$ m/s and $30$ m/s, corresponding to 
wind force scale $10$ and $11$, respectively, in the Beaufort 
scale~\cite{Singleton:2008}. 







\begin{thebibliography}{73}%
\makeatletter
\providecommand \@ifxundefined [1]{%
 \@ifx{#1\undefined}
}%
\providecommand \@ifnum [1]{%
 \ifnum #1\expandafter \@firstoftwo
 \else \expandafter \@secondoftwo
 \fi
}%
\providecommand \@ifx [1]{%
 \ifx #1\expandafter \@firstoftwo
 \else \expandafter \@secondoftwo
 \fi
}%
\providecommand \natexlab [1]{#1}%
\providecommand \enquote  [1]{``#1''}%
\providecommand \bibnamefont  [1]{#1}%
\providecommand \bibfnamefont [1]{#1}%
\providecommand \citenamefont [1]{#1}%
\providecommand \href@noop [0]{\@secondoftwo}%
\providecommand \href [0]{\begingroup \@sanitize@url \@href}%
\providecommand \@href[1]{\@@startlink{#1}\@@href}%
\providecommand \@@href[1]{\endgroup#1\@@endlink}%
\providecommand \@sanitize@url [0]{\catcode `\\12\catcode `\$12\catcode
  `\&12\catcode `\#12\catcode `\^12\catcode `\_12\catcode `\%12\relax}%
\providecommand \@@startlink[1]{}%
\providecommand \@@endlink[0]{}%
\providecommand \url  [0]{\begingroup\@sanitize@url \@url }%
\providecommand \@url [1]{\endgroup\@href {#1}{\urlprefix }}%
\providecommand \urlprefix  [0]{URL }%
\providecommand \Eprint [0]{\href }%
\providecommand \doibase [0]{https://doi.org/}%
\providecommand \selectlanguage [0]{\@gobble}%
\providecommand \bibinfo  [0]{\@secondoftwo}%
\providecommand \bibfield  [0]{\@secondoftwo}%
\providecommand \translation [1]{[#1]}%
\providecommand \BibitemOpen [0]{}%
\providecommand \bibitemStop [0]{}%
\providecommand \bibitemNoStop [0]{.\EOS\space}%
\providecommand \EOS [0]{\spacefactor3000\relax}%
\providecommand \BibitemShut  [1]{\csname bibitem#1\endcsname}%
\let\auto@bib@innerbib\@empty
\bibitem [{\citenamefont {Albeverio}\ \emph {et~al.}(2006)\citenamefont
  {Albeverio}, \citenamefont {Jentsch},\ and\ \citenamefont
  {Kantz}}]{AJK:book}%
  \BibitemOpen
  \bibfield  {author} {\bibinfo {author} {\bibfnamefont {S.}~\bibnamefont
  {Albeverio}}, \bibinfo {author} {\bibfnamefont {V.}~\bibnamefont {Jentsch}},\
  and\ \bibinfo {author} {\bibfnamefont {H.}~\bibnamefont {Kantz}},\
  }\href@noop {} {\emph {\bibinfo {title} {Extreme events in nature and
  society}}}\ (\bibinfo  {publisher} {Springer Science \& Business Media},\
  \bibinfo {year} {2006})\BibitemShut {NoStop}%
\bibitem [{\citenamefont {White}\ and\ \citenamefont
  {Fornberg}(1998)}]{WF:1998}%
  \BibitemOpen
  \bibfield  {author} {\bibinfo {author} {\bibfnamefont {B.~S.}\ \bibnamefont
  {White}}\ and\ \bibinfo {author} {\bibfnamefont {B.}~\bibnamefont
  {Fornberg}},\ }\bibfield  {title} {\bibinfo {title} {On the chance of freak
  waves at sea},\ }\href@noop {} {\bibfield  {journal} {\bibinfo  {journal} {J.
  Fluid Mech.}\ }\textbf {\bibinfo {volume} {355}},\ \bibinfo {pages} {113}
  (\bibinfo {year} {1998})}\BibitemShut {NoStop}%
\bibitem [{\citenamefont {Solli}\ \emph {et~al.}(2007)\citenamefont {Solli},
  \citenamefont {Ropers}, \citenamefont {Koonath},\ and\ \citenamefont
  {Jalali}}]{SRKB:2007}%
  \BibitemOpen
  \bibfield  {author} {\bibinfo {author} {\bibfnamefont {D.~R.}\ \bibnamefont
  {Solli}}, \bibinfo {author} {\bibfnamefont {C.}~\bibnamefont {Ropers}},
  \bibinfo {author} {\bibfnamefont {P.}~\bibnamefont {Koonath}},\ and\ \bibinfo
  {author} {\bibfnamefont {B.}~\bibnamefont {Jalali}},\ }\bibfield  {title}
  {\bibinfo {title} {Optical rogue waves},\ }\href@noop {} {\bibfield
  {journal} {\bibinfo  {journal} {Nature}\ }\textbf {\bibinfo {volume} {450}},\
  \bibinfo {pages} {1054} (\bibinfo {year} {2007})}\BibitemShut {NoStop}%
\bibitem [{\citenamefont {Akhmediev}\ \emph {et~al.}(2016)\citenamefont
  {Akhmediev}, \citenamefont {Kibler}, \citenamefont {Baronio}, \citenamefont
  {Beli\'{c}}, \citenamefont {Zhong}, \citenamefont {Zhang}, \citenamefont
  {Chang}, \citenamefont {Soto-Crespo}, \citenamefont {Vouzas},\ and\
  \citenamefont {Grelu}}]{Akhmedievetal:2016}%
  \BibitemOpen
  \bibfield  {author} {\bibinfo {author} {\bibfnamefont {N.}~\bibnamefont
  {Akhmediev}}, \bibinfo {author} {\bibfnamefont {B.}~\bibnamefont {Kibler}},
  \bibinfo {author} {\bibfnamefont {F.}~\bibnamefont {Baronio}}, \bibinfo
  {author} {\bibfnamefont {M.}~\bibnamefont {Beli\'{c}}}, \bibinfo {author}
  {\bibfnamefont {W.-P.}\ \bibnamefont {Zhong}}, \bibinfo {author}
  {\bibfnamefont {Y.-Q.}\ \bibnamefont {Zhang}}, \bibinfo {author}
  {\bibfnamefont {W.-K.}\ \bibnamefont {Chang}}, \bibinfo {author}
  {\bibfnamefont {J.~M.}\ \bibnamefont {Soto-Crespo}}, \bibinfo {author}
  {\bibfnamefont {P.}~\bibnamefont {Vouzas}},\ and\ \bibinfo {author}
  {\bibfnamefont {P.}~\bibnamefont {Grelu}},\ }\bibfield  {title} {\bibinfo
  {title} {Roadmap on optical rogue waves and extreme events},\ }\href@noop {}
  {\bibfield  {journal} {\bibinfo  {journal} {J. Opt.}\ }\textbf {\bibinfo
  {volume} {18}},\ \bibinfo {pages} {063001} (\bibinfo {year}
  {2016})}\BibitemShut {NoStop}%
\bibitem [{\citenamefont {Selmi}\ \emph {et~al.}(2016)\citenamefont {Selmi},
  \citenamefont {Coulibaly}, \citenamefont {Loghmari}, \citenamefont {Sagnes},
  \citenamefont {Beaudoin}, \citenamefont {Clerc},\ and\ \citenamefont
  {Barbay}}]{SCLSBCB:2016}%
  \BibitemOpen
  \bibfield  {author} {\bibinfo {author} {\bibfnamefont {F.}~\bibnamefont
  {Selmi}}, \bibinfo {author} {\bibfnamefont {S.}~\bibnamefont {Coulibaly}},
  \bibinfo {author} {\bibfnamefont {Z.}~\bibnamefont {Loghmari}}, \bibinfo
  {author} {\bibfnamefont {I.}~\bibnamefont {Sagnes}}, \bibinfo {author}
  {\bibfnamefont {G.}~\bibnamefont {Beaudoin}}, \bibinfo {author}
  {\bibfnamefont {M.~G.}\ \bibnamefont {Clerc}},\ and\ \bibinfo {author}
  {\bibfnamefont {S.}~\bibnamefont {Barbay}},\ }\bibfield  {title} {\bibinfo
  {title} {Spatiotemporal chaos induces extreme events in an extended
  microcavity laser},\ }\href {https://doi.org/10.1103/PhysRevLett.116.013901}
  {\bibfield  {journal} {\bibinfo  {journal} {Phys. Rev. Lett.}\ }\textbf
  {\bibinfo {volume} {116}},\ \bibinfo {pages} {013901} (\bibinfo {year}
  {2016})}\BibitemShut {NoStop}%
\bibitem [{\citenamefont {L’vov}\ \emph {et~al.}(2001)\citenamefont
  {L’vov}, \citenamefont {Pomyalov},\ and\ \citenamefont
  {Procaccia}}]{LPP:2001}%
  \BibitemOpen
  \bibfield  {author} {\bibinfo {author} {\bibfnamefont {V.~S.}\ \bibnamefont
  {L’vov}}, \bibinfo {author} {\bibfnamefont {A.}~\bibnamefont {Pomyalov}},\
  and\ \bibinfo {author} {\bibfnamefont {I.}~\bibnamefont {Procaccia}},\
  }\bibfield  {title} {\bibinfo {title} {Outliers, extreme events, and
  multiscaling},\ }\href@noop {} {\bibfield  {journal} {\bibinfo  {journal}
  {Phys. Rev. E}\ }\textbf {\bibinfo {volume} {63}},\ \bibinfo {pages} {056118}
  (\bibinfo {year} {2001})}\BibitemShut {NoStop}%
\bibitem [{\citenamefont {Santhanam}\ and\ \citenamefont
  {Kantz}(2005)}]{SK:2005}%
  \BibitemOpen
  \bibfield  {author} {\bibinfo {author} {\bibfnamefont {M.~S.}\ \bibnamefont
  {Santhanam}}\ and\ \bibinfo {author} {\bibfnamefont {H.}~\bibnamefont
  {Kantz}},\ }\bibfield  {title} {\bibinfo {title} {Long-range correlations and
  rare events in boundary layer wind fields},\ }\href@noop {} {\bibfield
  {journal} {\bibinfo  {journal} {Physica A}\ }\textbf {\bibinfo {volume}
  {345}},\ \bibinfo {pages} {713} (\bibinfo {year} {2005})}\BibitemShut
  {NoStop}%
\bibitem [{\citenamefont {Altmann}\ and\ \citenamefont
  {Kantz}(2005)}]{AK:2005}%
  \BibitemOpen
  \bibfield  {author} {\bibinfo {author} {\bibfnamefont {E.~G.}\ \bibnamefont
  {Altmann}}\ and\ \bibinfo {author} {\bibfnamefont {H.}~\bibnamefont
  {Kantz}},\ }\bibfield  {title} {\bibinfo {title} {Recurrence time analysis,
  long-term correlations, and extreme events},\ }\href@noop {} {\bibfield
  {journal} {\bibinfo  {journal} {Phys. Rev. E}\ }\textbf {\bibinfo {volume}
  {71}},\ \bibinfo {pages} {056106} (\bibinfo {year} {2005})}\BibitemShut
  {NoStop}%
\bibitem [{\citenamefont {Nicolis}\ \emph {et~al.}(2006)\citenamefont
  {Nicolis}, \citenamefont {Balakrishnan},\ and\ \citenamefont
  {Nicolis}}]{NBN:2006}%
  \BibitemOpen
  \bibfield  {author} {\bibinfo {author} {\bibfnamefont {C.}~\bibnamefont
  {Nicolis}}, \bibinfo {author} {\bibfnamefont {V.}~\bibnamefont
  {Balakrishnan}},\ and\ \bibinfo {author} {\bibfnamefont {G.}~\bibnamefont
  {Nicolis}},\ }\bibfield  {title} {\bibinfo {title} {Extreme events in
  deterministic dynamical systems},\ }\href
  {https://doi.org/10.1103/PhysRevLett.97.210602} {\bibfield  {journal}
  {\bibinfo  {journal} {Phys. Rev. Lett.}\ }\textbf {\bibinfo {volume} {97}},\
  \bibinfo {pages} {210602} (\bibinfo {year} {2006})}\BibitemShut {NoStop}%
\bibitem [{\citenamefont {Santhanam}\ and\ \citenamefont
  {Kantz}(2008)}]{SK:2008}%
  \BibitemOpen
  \bibfield  {author} {\bibinfo {author} {\bibfnamefont {M.}~\bibnamefont
  {Santhanam}}\ and\ \bibinfo {author} {\bibfnamefont {H.}~\bibnamefont
  {Kantz}},\ }\bibfield  {title} {\bibinfo {title} {Return interval
  distribution of extreme events and long-term memory},\ }\href@noop {}
  {\bibfield  {journal} {\bibinfo  {journal} {Phys. Rev. E}\ }\textbf {\bibinfo
  {volume} {78}},\ \bibinfo {pages} {051113} (\bibinfo {year}
  {2008})}\BibitemShut {NoStop}%
\bibitem [{\citenamefont {Ansmann}\ \emph {et~al.}(2013)\citenamefont
  {Ansmann}, \citenamefont {Karnatak}, \citenamefont {Lehnertz},\ and\
  \citenamefont {Feudel}}]{AKLF:2013}%
  \BibitemOpen
  \bibfield  {author} {\bibinfo {author} {\bibfnamefont {G.}~\bibnamefont
  {Ansmann}}, \bibinfo {author} {\bibfnamefont {R.}~\bibnamefont {Karnatak}},
  \bibinfo {author} {\bibfnamefont {K.}~\bibnamefont {Lehnertz}},\ and\
  \bibinfo {author} {\bibfnamefont {U.}~\bibnamefont {Feudel}},\ }\bibfield
  {title} {\bibinfo {title} {Extreme events in excitable systems and mechanisms
  of their generation},\ }\href {https://doi.org/10.1103/PhysRevE.88.052911}
  {\bibfield  {journal} {\bibinfo  {journal} {Phys. Rev. E}\ }\textbf {\bibinfo
  {volume} {88}},\ \bibinfo {pages} {052911} (\bibinfo {year}
  {2013})}\BibitemShut {NoStop}%
\bibitem [{\citenamefont {B\'odai}\ \emph {et~al.}(2013)\citenamefont
  {B\'odai}, \citenamefont {K\'arolyi},\ and\ \citenamefont
  {T\'el}}]{BKT:2013}%
  \BibitemOpen
  \bibfield  {author} {\bibinfo {author} {\bibfnamefont {T.}~\bibnamefont
  {B\'odai}}, \bibinfo {author} {\bibfnamefont {G.}~\bibnamefont {K\'arolyi}},\
  and\ \bibinfo {author} {\bibfnamefont {T.}~\bibnamefont {T\'el}},\ }\bibfield
   {title} {\bibinfo {title} {Driving a conceptual model climate by different
  processes: Snapshot attractors and extreme events},\ }\href
  {https://doi.org/10.1103/PhysRevE.87.022822} {\bibfield  {journal} {\bibinfo
  {journal} {Phys. Rev. E}\ }\textbf {\bibinfo {volume} {87}},\ \bibinfo
  {pages} {022822} (\bibinfo {year} {2013})}\BibitemShut {NoStop}%
\bibitem [{\citenamefont {Karnatak}\ \emph {et~al.}(2014)\citenamefont
  {Karnatak}, \citenamefont {Ansmann}, \citenamefont {Feudel},\ and\
  \citenamefont {Lehnertz}}]{KAFL:2014}%
  \BibitemOpen
  \bibfield  {author} {\bibinfo {author} {\bibfnamefont {R.}~\bibnamefont
  {Karnatak}}, \bibinfo {author} {\bibfnamefont {G.}~\bibnamefont {Ansmann}},
  \bibinfo {author} {\bibfnamefont {U.}~\bibnamefont {Feudel}},\ and\ \bibinfo
  {author} {\bibfnamefont {K.}~\bibnamefont {Lehnertz}},\ }\bibfield  {title}
  {\bibinfo {title} {Route to extreme events in excitable systems},\ }\href
  {https://doi.org/10.1103/PhysRevE.90.022917} {\bibfield  {journal} {\bibinfo
  {journal} {Phys. Rev. E}\ }\textbf {\bibinfo {volume} {90}},\ \bibinfo
  {pages} {022917} (\bibinfo {year} {2014})}\BibitemShut {NoStop}%
\bibitem [{\citenamefont {Mulhern}\ \emph {et~al.}(2015)\citenamefont
  {Mulhern}, \citenamefont {Bialonski},\ and\ \citenamefont
  {Kantz}}]{MBK:2015}%
  \BibitemOpen
  \bibfield  {author} {\bibinfo {author} {\bibfnamefont {C.}~\bibnamefont
  {Mulhern}}, \bibinfo {author} {\bibfnamefont {S.}~\bibnamefont {Bialonski}},\
  and\ \bibinfo {author} {\bibfnamefont {H.}~\bibnamefont {Kantz}},\ }\bibfield
   {title} {\bibinfo {title} {Extreme events due to localization of energy},\
  }\href {https://doi.org/10.1103/PhysRevE.91.012918} {\bibfield  {journal}
  {\bibinfo  {journal} {Phys. Rev. E}\ }\textbf {\bibinfo {volume} {91}},\
  \bibinfo {pages} {012918} (\bibinfo {year} {2015})}\BibitemShut {NoStop}%
\bibitem [{\citenamefont {Qi}\ and\ \citenamefont {Majda}(2016)}]{QM:2016}%
  \BibitemOpen
  \bibfield  {author} {\bibinfo {author} {\bibfnamefont {D.}~\bibnamefont
  {Qi}}\ and\ \bibinfo {author} {\bibfnamefont {A.~J.}\ \bibnamefont {Majda}},\
  }\bibfield  {title} {\bibinfo {title} {Predicting fat-tailed intermittent
  probability distributions in passive scalar turbulence with imperfect models
  through empirical information theory},\ }\href@noop {} {\bibfield  {journal}
  {\bibinfo  {journal} {Commun. Math. Sci.}\ }\textbf {\bibinfo {volume}
  {14}},\ \bibinfo {pages} {1687} (\bibinfo {year} {2016})}\BibitemShut
  {NoStop}%
\bibitem [{\citenamefont {Mohamad}\ and\ \citenamefont
  {Sapsis}(2018)}]{MS:2018}%
  \BibitemOpen
  \bibfield  {author} {\bibinfo {author} {\bibfnamefont {M.~A.}\ \bibnamefont
  {Mohamad}}\ and\ \bibinfo {author} {\bibfnamefont {T.~P.}\ \bibnamefont
  {Sapsis}},\ }\bibfield  {title} {\bibinfo {title} {Sequential sampling
  strategy for extreme event statistics in nonlinear dynamical systems},\
  }\href@noop {} {\bibfield  {journal} {\bibinfo  {journal} {Proc. Natl. Acad.
  Sci. (USA)}\ }\textbf {\bibinfo {volume} {115}},\ \bibinfo {pages} {11138}
  (\bibinfo {year} {2018})}\BibitemShut {NoStop}%
\bibitem [{\citenamefont {Bolles}\ \emph {et~al.}(2019)\citenamefont {Bolles},
  \citenamefont {Speer},\ and\ \citenamefont {Moore}}]{BSM:2019}%
  \BibitemOpen
  \bibfield  {author} {\bibinfo {author} {\bibfnamefont {C.~T.}\ \bibnamefont
  {Bolles}}, \bibinfo {author} {\bibfnamefont {K.}~\bibnamefont {Speer}},\ and\
  \bibinfo {author} {\bibfnamefont {M.~N.~J.}\ \bibnamefont {Moore}},\
  }\bibfield  {title} {\bibinfo {title} {Anomalous wave statistics induced by
  abrupt depth change},\ }\href
  {https://doi.org/10.1103/PhysRevFluids.4.011801} {\bibfield  {journal}
  {\bibinfo  {journal} {Phys. Rev. Fluids}\ }\textbf {\bibinfo {volume} {4}},\
  \bibinfo {pages} {011801} (\bibinfo {year} {2019})}\BibitemShut {NoStop}%
\bibitem [{\citenamefont {Hallerberg}\ \emph {et~al.}(2007)\citenamefont
  {Hallerberg}, \citenamefont {Altmann}, \citenamefont {Holstein},\ and\
  \citenamefont {Kantz}}]{HAHK:2007}%
  \BibitemOpen
  \bibfield  {author} {\bibinfo {author} {\bibfnamefont {S.}~\bibnamefont
  {Hallerberg}}, \bibinfo {author} {\bibfnamefont {E.~G.}\ \bibnamefont
  {Altmann}}, \bibinfo {author} {\bibfnamefont {D.}~\bibnamefont {Holstein}},\
  and\ \bibinfo {author} {\bibfnamefont {H.}~\bibnamefont {Kantz}},\ }\bibfield
   {title} {\bibinfo {title} {Precursors of extreme increments},\ }\href@noop
  {} {\bibfield  {journal} {\bibinfo  {journal} {Phys. Rev. E}\ }\textbf
  {\bibinfo {volume} {75}},\ \bibinfo {pages} {016706} (\bibinfo {year}
  {2007})}\BibitemShut {NoStop}%
\bibitem [{\citenamefont {Donner}\ and\ \citenamefont
  {Barbosa}(2008)}]{DB:2008}%
  \BibitemOpen
  \bibfield  {author} {\bibinfo {author} {\bibfnamefont {R.~V.}\ \bibnamefont
  {Donner}}\ and\ \bibinfo {author} {\bibfnamefont {S.~M.}\ \bibnamefont
  {Barbosa}},\ }\bibfield  {title} {\bibinfo {title} {Nonlinear time series
  analysis in the geosciences},\ }\href@noop {} {\bibfield  {journal} {\bibinfo
   {journal} {Lecture Notes in Earth Sciences}\ }\textbf {\bibinfo {volume}
  {112}} (\bibinfo {year} {2008})}\BibitemShut {NoStop}%
\bibitem [{\citenamefont {Cavalcante}\ \emph {et~al.}(2013)\citenamefont
  {Cavalcante}, \citenamefont {Ori{\'a}}, \citenamefont {Sornette},
  \citenamefont {Ott},\ and\ \citenamefont {Gauthier}}]{COSOG:2013}%
  \BibitemOpen
  \bibfield  {author} {\bibinfo {author} {\bibfnamefont {H.~L. d.~S.}\
  \bibnamefont {Cavalcante}}, \bibinfo {author} {\bibfnamefont
  {M.}~\bibnamefont {Ori{\'a}}}, \bibinfo {author} {\bibfnamefont
  {D.}~\bibnamefont {Sornette}}, \bibinfo {author} {\bibfnamefont
  {E.}~\bibnamefont {Ott}},\ and\ \bibinfo {author} {\bibfnamefont {D.~J.}\
  \bibnamefont {Gauthier}},\ }\bibfield  {title} {\bibinfo {title}
  {Predictability and suppression of extreme events in a chaotic system},\
  }\href@noop {} {\bibfield  {journal} {\bibinfo  {journal} {Phys. Rev. Lett.}\
  }\textbf {\bibinfo {volume} {111}},\ \bibinfo {pages} {198701} (\bibinfo
  {year} {2013})}\BibitemShut {NoStop}%
\bibitem [{\citenamefont {B{\'o}dai}\ and\ \citenamefont
  {Franzke}(2017)}]{BF:2017}%
  \BibitemOpen
  \bibfield  {author} {\bibinfo {author} {\bibfnamefont {T.}~\bibnamefont
  {B{\'o}dai}}\ and\ \bibinfo {author} {\bibfnamefont {C.}~\bibnamefont
  {Franzke}},\ }\bibfield  {title} {\bibinfo {title} {Predictability of
  fat-tailed extremes},\ }\href@noop {} {\bibfield  {journal} {\bibinfo
  {journal} {Phys. Rev. E}\ }\textbf {\bibinfo {volume} {96}},\ \bibinfo
  {pages} {032120} (\bibinfo {year} {2017})}\BibitemShut {NoStop}%
\bibitem [{\citenamefont {Nagy}\ and\ \citenamefont {Ott}(2007)}]{NO:2007}%
  \BibitemOpen
  \bibfield  {author} {\bibinfo {author} {\bibfnamefont {V.}~\bibnamefont
  {Nagy}}\ and\ \bibinfo {author} {\bibfnamefont {E.}~\bibnamefont {Ott}},\
  }\bibfield  {title} {\bibinfo {title} {Control of rare intense events in
  spatiotemporally chaotic systems},\ }\href
  {https://doi.org/10.1103/PhysRevE.76.066206} {\bibfield  {journal} {\bibinfo
  {journal} {Phys. Rev. E}\ }\textbf {\bibinfo {volume} {76}},\ \bibinfo
  {pages} {066206} (\bibinfo {year} {2007})}\BibitemShut {NoStop}%
\bibitem [{\citenamefont {Du}\ \emph {et~al.}(2008)\citenamefont {Du},
  \citenamefont {Chen}, \citenamefont {Lai},\ and\ \citenamefont
  {Xu}}]{DCLX:2008}%
  \BibitemOpen
  \bibfield  {author} {\bibinfo {author} {\bibfnamefont {L.}~\bibnamefont
  {Du}}, \bibinfo {author} {\bibfnamefont {Q.}~\bibnamefont {Chen}}, \bibinfo
  {author} {\bibfnamefont {Y.-C.}\ \bibnamefont {Lai}},\ and\ \bibinfo {author}
  {\bibfnamefont {W.}~\bibnamefont {Xu}},\ }\bibfield  {title} {\bibinfo
  {title} {Observation-based control of rare intense events in the complex
  {Ginzburg-Landau} equation},\ }\href@noop {} {\bibfield  {journal} {\bibinfo
  {journal} {Phys. Rev. E}\ }\textbf {\bibinfo {volume} {78}},\ \bibinfo
  {pages} {015201} (\bibinfo {year} {2008})}\BibitemShut {NoStop}%
\bibitem [{\citenamefont {Bialonski}\ \emph {et~al.}(2015)\citenamefont
  {Bialonski}, \citenamefont {Ansmann},\ and\ \citenamefont
  {Kantz}}]{BAK:2015}%
  \BibitemOpen
  \bibfield  {author} {\bibinfo {author} {\bibfnamefont {S.}~\bibnamefont
  {Bialonski}}, \bibinfo {author} {\bibfnamefont {G.}~\bibnamefont {Ansmann}},\
  and\ \bibinfo {author} {\bibfnamefont {H.}~\bibnamefont {Kantz}},\ }\bibfield
   {title} {\bibinfo {title} {Data-driven prediction and prevention of extreme
  events in a spatially extended excitable system},\ }\href
  {https://doi.org/10.1103/PhysRevE.92.042910} {\bibfield  {journal} {\bibinfo
  {journal} {Phys. Rev. E}\ }\textbf {\bibinfo {volume} {92}},\ \bibinfo
  {pages} {042910} (\bibinfo {year} {2015})}\BibitemShut {NoStop}%
\bibitem [{\citenamefont {Lopatka}(2019)}]{Lopatka:2019}%
  \BibitemOpen
  \bibfield  {author} {\bibinfo {author} {\bibfnamefont {A.}~\bibnamefont
  {Lopatka}},\ }\bibfield  {title} {\bibinfo {title} {Meteorologists predict
  better weather forecasting with {AI}},\ }\href@noop {} {\bibfield  {journal}
  {\bibinfo  {journal} {Phys. Today}\ }\textbf {\bibinfo {volume} {72}},\
  \bibinfo {pages} {32} (\bibinfo {year} {2019})}\BibitemShut {NoStop}%
\bibitem [{\citenamefont {Hansen}\ \emph {et~al.}(2000)\citenamefont {Hansen},
  \citenamefont {Sato}, \citenamefont {Ruedy}, \citenamefont {Lacis},\ and\
  \citenamefont {Oinas}}]{HSRLO:2000}%
  \BibitemOpen
  \bibfield  {author} {\bibinfo {author} {\bibfnamefont {J.}~\bibnamefont
  {Hansen}}, \bibinfo {author} {\bibfnamefont {M.}~\bibnamefont {Sato}},
  \bibinfo {author} {\bibfnamefont {R.}~\bibnamefont {Ruedy}}, \bibinfo
  {author} {\bibfnamefont {A.}~\bibnamefont {Lacis}},\ and\ \bibinfo {author}
  {\bibfnamefont {V.}~\bibnamefont {Oinas}},\ }\bibfield  {title} {\bibinfo
  {title} {Global warming in the twenty-first century: An alternative
  scenario},\ }\href@noop {} {\bibfield  {journal} {\bibinfo  {journal} {Proc.
  Natl. Acad. Sci. (USA)}\ }\textbf {\bibinfo {volume} {97}},\ \bibinfo {pages}
  {9875} (\bibinfo {year} {2000})}\BibitemShut {NoStop}%
\bibitem [{\citenamefont {Easterling}\ \emph {et~al.}(2000)\citenamefont
  {Easterling}, \citenamefont {Evans}, \citenamefont {Groisman}, \citenamefont
  {Karl}, \citenamefont {Kunkel},\ and\ \citenamefont {Ambenje}}]{EEGKKA:2000}%
  \BibitemOpen
  \bibfield  {author} {\bibinfo {author} {\bibfnamefont {D.~R.}\ \bibnamefont
  {Easterling}}, \bibinfo {author} {\bibfnamefont {J.}~\bibnamefont {Evans}},
  \bibinfo {author} {\bibfnamefont {P.~Y.}\ \bibnamefont {Groisman}}, \bibinfo
  {author} {\bibfnamefont {T.~R.}\ \bibnamefont {Karl}}, \bibinfo {author}
  {\bibfnamefont {K.~E.}\ \bibnamefont {Kunkel}},\ and\ \bibinfo {author}
  {\bibfnamefont {P.}~\bibnamefont {Ambenje}},\ }\bibfield  {title} {\bibinfo
  {title} {Observed variability and trends in extreme climate events: a brief
  review},\ }\href@noop {} {\bibfield  {journal} {\bibinfo  {journal} {Bull.
  Ame. Meteorol. Soc.}\ }\textbf {\bibinfo {volume} {81}},\ \bibinfo {pages}
  {417} (\bibinfo {year} {2000})}\BibitemShut {NoStop}%
\bibitem [{\citenamefont {Palmer}\ and\ \citenamefont
  {R{\"a}is{\"a}nen}(2002)}]{PR:2002}%
  \BibitemOpen
  \bibfield  {author} {\bibinfo {author} {\bibfnamefont {T.}~\bibnamefont
  {Palmer}}\ and\ \bibinfo {author} {\bibfnamefont {J.}~\bibnamefont
  {R{\"a}is{\"a}nen}},\ }\bibfield  {title} {\bibinfo {title} {Quantifying the
  risk of extreme seasonal precipitation events in a changing climate},\
  }\href@noop {} {\bibfield  {journal} {\bibinfo  {journal} {Nature}\ }\textbf
  {\bibinfo {volume} {415}},\ \bibinfo {pages} {512} (\bibinfo {year}
  {2002})}\BibitemShut {NoStop}%
\bibitem [{\citenamefont {Guardian}(2012)}]{CC:2012}%
  \BibitemOpen
  \bibfield  {author} {\bibinfo {author} {\bibfnamefont {T.}~\bibnamefont
  {Guardian}},\ }\bibfield  {title} {\bibinfo {title} {Scientists attribute
  extreme weather to man-made climate change},\ }\href@noop {} {\bibfield
  {journal} {\bibinfo  {journal}
  {https://www.theguardian.com/environment/2012/jul/10/extreme-weather-manmade-climate-change}\
  } (\bibinfo {year} {2012})}\BibitemShut {NoStop}%
\bibitem [{\citenamefont {for Environmental~Information}(2016)}]{NOAA:2016}%
  \BibitemOpen
  \bibfield  {author} {\bibinfo {author} {\bibfnamefont {N.~N.~C.}\
  \bibnamefont {for Environmental~Information}},\ }\bibfield  {title} {\bibinfo
  {title} {Billion-dollar weather and climate disasters: Summary stats},\
  }\href@noop {} {\bibfield  {journal} {\bibinfo  {journal}
  {http://www.ncdc.noaa.gov/billions/summary-stats}\ } (\bibinfo {year}
  {2016})}\BibitemShut {NoStop}%
\bibitem [{\citenamefont {Stott}(2016)}]{Stott:2016}%
  \BibitemOpen
  \bibfield  {author} {\bibinfo {author} {\bibfnamefont {P.}~\bibnamefont
  {Stott}},\ }\bibfield  {title} {\bibinfo {title} {How climate change affects
  extreme weather events},\ }\href@noop {} {\bibfield  {journal} {\bibinfo
  {journal} {Science}\ }\textbf {\bibinfo {volume} {352}},\ \bibinfo {pages}
  {1517} (\bibinfo {year} {2016})}\BibitemShut {NoStop}%
\bibitem [{\citenamefont {Qi}\ and\ \citenamefont {Majda}(2018)}]{QM:2018}%
  \BibitemOpen
  \bibfield  {author} {\bibinfo {author} {\bibfnamefont {D.}~\bibnamefont
  {Qi}}\ and\ \bibinfo {author} {\bibfnamefont {A.~J.}\ \bibnamefont {Majda}},\
  }\bibfield  {title} {\bibinfo {title} {Predicting extreme events for passive
  scalar turbulence in two-layer baroclinic flows through reduced-order
  stochastic models},\ }\href@noop {} {\bibfield  {journal} {\bibinfo
  {journal} {Commun. Math. Sci.}\ }\textbf {\bibinfo {volume} {16}},\ \bibinfo
  {pages} {17} (\bibinfo {year} {2018})}\BibitemShut {NoStop}%
\bibitem [{\citenamefont {Majda}\ \emph {et~al.}(2019)\citenamefont {Majda},
  \citenamefont {Moore},\ and\ \citenamefont {Qi}}]{MMQ:2019}%
  \BibitemOpen
  \bibfield  {author} {\bibinfo {author} {\bibfnamefont {A.~J.}\ \bibnamefont
  {Majda}}, \bibinfo {author} {\bibfnamefont {M.~N.~J.}\ \bibnamefont
  {Moore}},\ and\ \bibinfo {author} {\bibfnamefont {D.}~\bibnamefont {Qi}},\
  }\bibfield  {title} {\bibinfo {title} {Statistical dynamical model to predict
  extreme events and anomalous features in shallow water waves with abrupt
  depth change},\ }\href@noop {} {\bibfield  {journal} {\bibinfo  {journal}
  {Proc. Natl. Acad. Sci. (USA)}\ }\textbf {\bibinfo {volume} {116}},\ \bibinfo
  {pages} {3982} (\bibinfo {year} {2019})}\BibitemShut {NoStop}%
\bibitem [{\citenamefont {LeCun}\ \emph {et~al.}(2015)\citenamefont {LeCun},
  \citenamefont {Bengio},\ and\ \citenamefont {Hinton}}]{LBH:2015}%
  \BibitemOpen
  \bibfield  {author} {\bibinfo {author} {\bibfnamefont {Y.}~\bibnamefont
  {LeCun}}, \bibinfo {author} {\bibfnamefont {Y.}~\bibnamefont {Bengio}},\ and\
  \bibinfo {author} {\bibfnamefont {G.}~\bibnamefont {Hinton}},\ }\bibfield
  {title} {\bibinfo {title} {Deep learning},\ }\href@noop {} {\bibfield
  {journal} {\bibinfo  {journal} {Nature}\ }\textbf {\bibinfo {volume} {521}},\
  \bibinfo {pages} {436} (\bibinfo {year} {2015})}\BibitemShut {NoStop}%
\bibitem [{\citenamefont {Jordan}\ and\ \citenamefont
  {Mitchell}(2015)}]{JM:2015}%
  \BibitemOpen
  \bibfield  {author} {\bibinfo {author} {\bibfnamefont {M.~I.}\ \bibnamefont
  {Jordan}}\ and\ \bibinfo {author} {\bibfnamefont {T.~M.}\ \bibnamefont
  {Mitchell}},\ }\bibfield  {title} {\bibinfo {title} {Machine learning:
  Trends, perspectives, and prospects},\ }\href@noop {} {\bibfield  {journal}
  {\bibinfo  {journal} {Science}\ }\textbf {\bibinfo {volume} {349}},\ \bibinfo
  {pages} {255} (\bibinfo {year} {2015})}\BibitemShut {NoStop}%
\bibitem [{\citenamefont {Goodfellow}\ \emph {et~al.}(2016)\citenamefont
  {Goodfellow}, \citenamefont {Bengio},\ and\ \citenamefont
  {Courville}}]{GBC:book}%
  \BibitemOpen
  \bibfield  {author} {\bibinfo {author} {\bibfnamefont {I.}~\bibnamefont
  {Goodfellow}}, \bibinfo {author} {\bibfnamefont {Y.}~\bibnamefont {Bengio}},\
  and\ \bibinfo {author} {\bibfnamefont {A.}~\bibnamefont {Courville}},\
  }\href@noop {} {\emph {\bibinfo {title} {Deep Learning}}}\ (\bibinfo
  {publisher} {The MIT Press},\ \bibinfo {address} {Cambridge, Massachusetts},\
  \bibinfo {year} {2016})\BibitemShut {NoStop}%
\bibitem [{\citenamefont {He}\ \emph {et~al.}(2016)\citenamefont {He},
  \citenamefont {Zhang}, \citenamefont {Ren},\ and\ \citenamefont
  {Sun}}]{HZRS:2016}%
  \BibitemOpen
  \bibfield  {author} {\bibinfo {author} {\bibfnamefont {K.}~\bibnamefont
  {He}}, \bibinfo {author} {\bibfnamefont {X.}~\bibnamefont {Zhang}}, \bibinfo
  {author} {\bibfnamefont {S.}~\bibnamefont {Ren}},\ and\ \bibinfo {author}
  {\bibfnamefont {J.}~\bibnamefont {Sun}},\ }\bibfield  {title} {\bibinfo
  {title} {Deep residual learning for image recognition},\ }in\ \href@noop {}
  {\emph {\bibinfo {booktitle} {CVPR}}}\ (\bibinfo {year} {2016})\ pp.\
  \bibinfo {pages} {770--778}\BibitemShut {NoStop}%
\bibitem [{\citenamefont {Krizhevsky}\ \emph {et~al.}(2012)\citenamefont
  {Krizhevsky}, \citenamefont {Sutskever},\ and\ \citenamefont
  {Hinton}}]{KSH:2012}%
  \BibitemOpen
  \bibfield  {author} {\bibinfo {author} {\bibfnamefont {A.}~\bibnamefont
  {Krizhevsky}}, \bibinfo {author} {\bibfnamefont {I.}~\bibnamefont
  {Sutskever}},\ and\ \bibinfo {author} {\bibfnamefont {G.~E.}\ \bibnamefont
  {Hinton}},\ }\bibfield  {title} {\bibinfo {title} {Imagenet classification
  with deep convolutional neural networks},\ }in\ \href@noop {} {\emph
  {\bibinfo {booktitle} {NeurIPS}}}\ (\bibinfo {year} {2012})\ pp.\ \bibinfo
  {pages} {1097--1105}\BibitemShut {NoStop}%
\bibitem [{\citenamefont {Hinton}\ \emph {et~al.}(2012)\citenamefont {Hinton},
  \citenamefont {Deng}, \citenamefont {Yu}, \citenamefont {Dahl}, \citenamefont
  {Mohamed}, \citenamefont {Jaitly}, \citenamefont {Senior}, \citenamefont
  {Vanhoucke}, \citenamefont {Nguyen}, \citenamefont {Kingsbury} \emph
  {et~al.}}]{HDYDMJSVNK:2012}%
  \BibitemOpen
  \bibfield  {author} {\bibinfo {author} {\bibfnamefont {G.}~\bibnamefont
  {Hinton}}, \bibinfo {author} {\bibfnamefont {L.}~\bibnamefont {Deng}},
  \bibinfo {author} {\bibfnamefont {D.}~\bibnamefont {Yu}}, \bibinfo {author}
  {\bibfnamefont {G.}~\bibnamefont {Dahl}}, \bibinfo {author} {\bibfnamefont
  {A.-r.}\ \bibnamefont {Mohamed}}, \bibinfo {author} {\bibfnamefont
  {N.}~\bibnamefont {Jaitly}}, \bibinfo {author} {\bibfnamefont
  {A.}~\bibnamefont {Senior}}, \bibinfo {author} {\bibfnamefont
  {V.}~\bibnamefont {Vanhoucke}}, \bibinfo {author} {\bibfnamefont
  {P.}~\bibnamefont {Nguyen}}, \bibinfo {author} {\bibfnamefont
  {B.}~\bibnamefont {Kingsbury}}, \emph {et~al.},\ }\bibfield  {title}
  {\bibinfo {title} {Deep neural networks for acoustic modeling in speech
  recognition},\ }\href@noop {} {\bibfield  {journal} {\bibinfo  {journal}
  {IEEE Signal Process. Mag.}\ }\textbf {\bibinfo {volume} {29}} (\bibinfo
  {year} {2012})}\BibitemShut {NoStop}%
\bibitem [{\citenamefont {Ma}\ \emph {et~al.}(2015)\citenamefont {Ma},
  \citenamefont {Sheridan}, \citenamefont {Liaw}, \citenamefont {Dahl},\ and\
  \citenamefont {Svetnik}}]{MSLDS:2015}%
  \BibitemOpen
  \bibfield  {author} {\bibinfo {author} {\bibfnamefont {J.}~\bibnamefont
  {Ma}}, \bibinfo {author} {\bibfnamefont {R.~P.}\ \bibnamefont {Sheridan}},
  \bibinfo {author} {\bibfnamefont {A.}~\bibnamefont {Liaw}}, \bibinfo {author}
  {\bibfnamefont {G.~E.}\ \bibnamefont {Dahl}},\ and\ \bibinfo {author}
  {\bibfnamefont {V.}~\bibnamefont {Svetnik}},\ }\bibfield  {title} {\bibinfo
  {title} {Deep neural nets as a method for quantitative structure--activity
  relationships},\ }\href@noop {} {\bibfield  {journal} {\bibinfo  {journal}
  {J. Chem. Inf. Model.}\ }\textbf {\bibinfo {volume} {55}},\ \bibinfo {pages}
  {263} (\bibinfo {year} {2015})}\BibitemShut {NoStop}%
\bibitem [{\citenamefont {Dang}\ \emph {et~al.}(2020)\citenamefont {Dang},
  \citenamefont {Gao}, \citenamefont {Sun}, \citenamefont {Li}, \citenamefont
  {Cai},\ and\ \citenamefont {Grebogi}}]{DGSLCG:2020}%
  \BibitemOpen
  \bibfield  {author} {\bibinfo {author} {\bibfnamefont {W.}~\bibnamefont
  {Dang}}, \bibinfo {author} {\bibfnamefont {Z.}~\bibnamefont {Gao}}, \bibinfo
  {author} {\bibfnamefont {X.}~\bibnamefont {Sun}}, \bibinfo {author}
  {\bibfnamefont {R.~L.}\ \bibnamefont {Li}}, \bibinfo {author} {\bibfnamefont
  {Q.}~\bibnamefont {Cai}},\ and\ \bibinfo {author} {\bibfnamefont
  {C.}~\bibnamefont {Grebogi}},\ }\bibfield  {title} {\bibinfo {title}
  {Multilayer brain network combined with deep convolutional neural network for
  detecting major depressive disorder},\ }\href@noop {} {\bibfield  {journal}
  {\bibinfo  {journal} {Nonlin. Dyn.}\ }\textbf {\bibinfo {volume} {102}},\
  \bibinfo {pages} {667} (\bibinfo {year} {2020})}\BibitemShut {NoStop}%
\bibitem [{\citenamefont {Silver}\ \emph {et~al.}(2017)\citenamefont {Silver},
  \citenamefont {Schrittwieser}, \citenamefont {Simonyan}, \citenamefont
  {Antonoglou}, \citenamefont {Huang}, \citenamefont {Guez}, \citenamefont
  {Hubert}, \citenamefont {Baker}, \citenamefont {Lai}, \citenamefont {Bolton}
  \emph {et~al.}}]{SSSAHGHBLB:2017}%
  \BibitemOpen
  \bibfield  {author} {\bibinfo {author} {\bibfnamefont {D.}~\bibnamefont
  {Silver}}, \bibinfo {author} {\bibfnamefont {J.}~\bibnamefont
  {Schrittwieser}}, \bibinfo {author} {\bibfnamefont {K.}~\bibnamefont
  {Simonyan}}, \bibinfo {author} {\bibfnamefont {I.}~\bibnamefont
  {Antonoglou}}, \bibinfo {author} {\bibfnamefont {A.}~\bibnamefont {Huang}},
  \bibinfo {author} {\bibfnamefont {A.}~\bibnamefont {Guez}}, \bibinfo {author}
  {\bibfnamefont {T.}~\bibnamefont {Hubert}}, \bibinfo {author} {\bibfnamefont
  {L.}~\bibnamefont {Baker}}, \bibinfo {author} {\bibfnamefont
  {M.}~\bibnamefont {Lai}}, \bibinfo {author} {\bibfnamefont {A.}~\bibnamefont
  {Bolton}}, \emph {et~al.},\ }\bibfield  {title} {\bibinfo {title} {Mastering
  the game of go without human knowledge},\ }\href@noop {} {\bibfield
  {journal} {\bibinfo  {journal} {Nature}\ }\textbf {\bibinfo {volume} {550}},\
  \bibinfo {pages} {354} (\bibinfo {year} {2017})}\BibitemShut {NoStop}%
\bibitem [{\citenamefont {Reichstein}\ \emph {et~al.}(2019)\citenamefont
  {Reichstein}, \citenamefont {Camps-Valls}, \citenamefont {Stevens},
  \citenamefont {Jung}, \citenamefont {Denzler}, \citenamefont {Carvalhais}
  \emph {et~al.}}]{RCSJDC:2019}%
  \BibitemOpen
  \bibfield  {author} {\bibinfo {author} {\bibfnamefont {M.}~\bibnamefont
  {Reichstein}}, \bibinfo {author} {\bibfnamefont {G.}~\bibnamefont
  {Camps-Valls}}, \bibinfo {author} {\bibfnamefont {B.}~\bibnamefont
  {Stevens}}, \bibinfo {author} {\bibfnamefont {M.}~\bibnamefont {Jung}},
  \bibinfo {author} {\bibfnamefont {J.}~\bibnamefont {Denzler}}, \bibinfo
  {author} {\bibfnamefont {N.}~\bibnamefont {Carvalhais}}, \emph {et~al.},\
  }\bibfield  {title} {\bibinfo {title} {Deep learning and process
  understanding for data-driven earth system science},\ }\href@noop {}
  {\bibfield  {journal} {\bibinfo  {journal} {Nature}\ }\textbf {\bibinfo
  {volume} {566}},\ \bibinfo {pages} {195} (\bibinfo {year}
  {2019})}\BibitemShut {NoStop}%
\bibitem [{\citenamefont {Ham}\ \emph {et~al.}(2019)\citenamefont {Ham},
  \citenamefont {Kim},\ and\ \citenamefont {Luo}}]{HKL:2019}%
  \BibitemOpen
  \bibfield  {author} {\bibinfo {author} {\bibfnamefont {Y.-G.}\ \bibnamefont
  {Ham}}, \bibinfo {author} {\bibfnamefont {J.-H.}\ \bibnamefont {Kim}},\ and\
  \bibinfo {author} {\bibfnamefont {J.-J.}\ \bibnamefont {Luo}},\ }\bibfield
  {title} {\bibinfo {title} {Deep learning for multi-year {ENSO} forecasts},\
  }\href@noop {} {\bibfield  {journal} {\bibinfo  {journal} {Nature}\ ,\
  \bibinfo {pages} {1}} (\bibinfo {year} {2019})}\BibitemShut {NoStop}%
\bibitem [{\citenamefont {Guth}\ and\ \citenamefont {Sapsis}(2019)}]{GS:2019}%
  \BibitemOpen
  \bibfield  {author} {\bibinfo {author} {\bibfnamefont {S.}~\bibnamefont
  {Guth}}\ and\ \bibinfo {author} {\bibfnamefont {T.~P.}\ \bibnamefont
  {Sapsis}},\ }\bibfield  {title} {\bibinfo {title} {Machine learning
  predictors of extreme events occurring in complex dynamical systems},\
  }\href@noop {} {\bibfield  {journal} {\bibinfo  {journal} {Entropy}\ }\textbf
  {\bibinfo {volume} {21}},\ \bibinfo {pages} {925} (\bibinfo {year}
  {2019})}\BibitemShut {NoStop}%
\bibitem [{\citenamefont {Qi}\ and\ \citenamefont {Majda}(2020)}]{QM:2020}%
  \BibitemOpen
  \bibfield  {author} {\bibinfo {author} {\bibfnamefont {D.}~\bibnamefont
  {Qi}}\ and\ \bibinfo {author} {\bibfnamefont {A.~J.}\ \bibnamefont {Majda}},\
  }\bibfield  {title} {\bibinfo {title} {Using machine learning to predict
  extreme events in complex systems},\ }\href@noop {} {\bibfield  {journal}
  {\bibinfo  {journal} {Proc. Natl. Acad. Sci. (USA)}\ }\textbf {\bibinfo
  {volume} {117}},\ \bibinfo {pages} {52} (\bibinfo {year} {2020})}\BibitemShut
  {NoStop}%
\bibitem [{\citenamefont {LeCun}\ \emph {et~al.}(1998)\citenamefont {LeCun},
  \citenamefont {Bottou}, \citenamefont {Bengio}, \citenamefont {Haffner} \emph
  {et~al.}}]{LBBH:1998}%
  \BibitemOpen
  \bibfield  {author} {\bibinfo {author} {\bibfnamefont {Y.}~\bibnamefont
  {LeCun}}, \bibinfo {author} {\bibfnamefont {L.}~\bibnamefont {Bottou}},
  \bibinfo {author} {\bibfnamefont {Y.}~\bibnamefont {Bengio}}, \bibinfo
  {author} {\bibfnamefont {P.}~\bibnamefont {Haffner}}, \emph {et~al.},\
  }\bibfield  {title} {\bibinfo {title} {Gradient-based learning applied to
  document recognition},\ }\href@noop {} {\bibfield  {journal} {\bibinfo
  {journal} {Proc. IEEE}\ }\textbf {\bibinfo {volume} {86}},\ \bibinfo {pages}
  {2278} (\bibinfo {year} {1998})}\BibitemShut {NoStop}%
\bibitem [{\citenamefont {Ciresan}\ \emph {et~al.}(2012)\citenamefont
  {Ciresan}, \citenamefont {Giusti}, \citenamefont {Gambardella},\ and\
  \citenamefont {Schmidhuber}}]{CGGS:2012}%
  \BibitemOpen
  \bibfield  {author} {\bibinfo {author} {\bibfnamefont {D.}~\bibnamefont
  {Ciresan}}, \bibinfo {author} {\bibfnamefont {A.}~\bibnamefont {Giusti}},
  \bibinfo {author} {\bibfnamefont {L.~M.}\ \bibnamefont {Gambardella}},\ and\
  \bibinfo {author} {\bibfnamefont {J.}~\bibnamefont {Schmidhuber}},\
  }\bibfield  {title} {\bibinfo {title} {Deep neural networks segment neuronal
  membranes in electron microscopy images},\ }in\ \href@noop {} {\emph
  {\bibinfo {booktitle} {NeurIPS}}}\ (\bibinfo {year} {2012})\ pp.\ \bibinfo
  {pages} {2843--2851}\BibitemShut {NoStop}%
\bibitem [{\citenamefont {Farabet}\ \emph {et~al.}(2012)\citenamefont
  {Farabet}, \citenamefont {Couprie}, \citenamefont {Najman},\ and\
  \citenamefont {LeCun}}]{FCNL:2012}%
  \BibitemOpen
  \bibfield  {author} {\bibinfo {author} {\bibfnamefont {C.}~\bibnamefont
  {Farabet}}, \bibinfo {author} {\bibfnamefont {C.}~\bibnamefont {Couprie}},
  \bibinfo {author} {\bibfnamefont {L.}~\bibnamefont {Najman}},\ and\ \bibinfo
  {author} {\bibfnamefont {Y.}~\bibnamefont {LeCun}},\ }\bibfield  {title}
  {\bibinfo {title} {Learning hierarchical features for scene labeling},\
  }\href@noop {} {\bibfield  {journal} {\bibinfo  {journal} {IEEE Trans.
  Pattern Anal. Mach. Intell.}\ }\textbf {\bibinfo {volume} {35}},\ \bibinfo
  {pages} {1915} (\bibinfo {year} {2012})}\BibitemShut {NoStop}%
\bibitem [{\citenamefont {Aranson}\ and\ \citenamefont
  {Kramer}(2002)}]{AK:2002}%
  \BibitemOpen
  \bibfield  {author} {\bibinfo {author} {\bibfnamefont {I.~S.}\ \bibnamefont
  {Aranson}}\ and\ \bibinfo {author} {\bibfnamefont {L.}~\bibnamefont
  {Kramer}},\ }\bibfield  {title} {\bibinfo {title} {The world of the complex
  {Ginzburg-Landau} equation},\ }\href@noop {} {\bibfield  {journal} {\bibinfo
  {journal} {Rev. Mod. Phys}\ }\textbf {\bibinfo {volume} {74}},\ \bibinfo
  {pages} {99} (\bibinfo {year} {2002})}\BibitemShut {NoStop}%
\bibitem [{\citenamefont {Cross}\ and\ \citenamefont
  {Hohenberg}(1993)}]{CH:1993}%
  \BibitemOpen
  \bibfield  {author} {\bibinfo {author} {\bibfnamefont {M.~C.}\ \bibnamefont
  {Cross}}\ and\ \bibinfo {author} {\bibfnamefont {P.~C.}\ \bibnamefont
  {Hohenberg}},\ }\bibfield  {title} {\bibinfo {title} {Pattern formation
  outside of equilibrium},\ }\href {https://doi.org/10.1103/RevModPhys.65.851}
  {\bibfield  {journal} {\bibinfo  {journal} {Rev. Mod. Phys.}\ }\textbf
  {\bibinfo {volume} {65}},\ \bibinfo {pages} {851} (\bibinfo {year}
  {1993})}\BibitemShut {NoStop}%
\bibitem [{\citenamefont {Kuramoto}(1984)}]{Kbook:1984}%
  \BibitemOpen
  \bibfield  {author} {\bibinfo {author} {\bibfnamefont {Y.}~\bibnamefont
  {Kuramoto}},\ }\href@noop {} {\emph {\bibinfo {title} {Chemical Oscillations,
  Waves and Turbulence}}}\ (\bibinfo  {publisher} {Springer},\ \bibinfo
  {address} {Berlin},\ \bibinfo {year} {1984})\BibitemShut {NoStop}%
\bibitem [{\citenamefont {Paszke}\ \emph {et~al.}(2017)\citenamefont {Paszke},
  \citenamefont {Gross}, \citenamefont {Chintala}, \citenamefont {Chanan},
  \citenamefont {Yang}, \citenamefont {DeVito}, \citenamefont {Lin},
  \citenamefont {Desmaison}, \citenamefont {Antiga},\ and\ \citenamefont
  {Lerer}}]{PGCCYDLDAL:2017}%
  \BibitemOpen
  \bibfield  {author} {\bibinfo {author} {\bibfnamefont {A.}~\bibnamefont
  {Paszke}}, \bibinfo {author} {\bibfnamefont {S.}~\bibnamefont {Gross}},
  \bibinfo {author} {\bibfnamefont {S.}~\bibnamefont {Chintala}}, \bibinfo
  {author} {\bibfnamefont {G.}~\bibnamefont {Chanan}}, \bibinfo {author}
  {\bibfnamefont {E.}~\bibnamefont {Yang}}, \bibinfo {author} {\bibfnamefont
  {Z.}~\bibnamefont {DeVito}}, \bibinfo {author} {\bibfnamefont
  {Z.}~\bibnamefont {Lin}}, \bibinfo {author} {\bibfnamefont {A.}~\bibnamefont
  {Desmaison}}, \bibinfo {author} {\bibfnamefont {L.}~\bibnamefont {Antiga}},\
  and\ \bibinfo {author} {\bibfnamefont {A.}~\bibnamefont {Lerer}},\ }\bibfield
   {title} {\bibinfo {title} {Automatic differentiation in pytorch},\ }in\
  \href@noop {} {\emph {\bibinfo {booktitle} {NIPS-W}}}\ (\bibinfo {year}
  {2017})\BibitemShut {NoStop}%
\bibitem [{\citenamefont {Paszke}\ \emph {et~al.}(2019)\citenamefont {Paszke},
  \citenamefont {Gross}, \citenamefont {Massa}, \citenamefont {Lerer},
  \citenamefont {Bradbury}, \citenamefont {Chanan}, \citenamefont {Killeen},
  \citenamefont {Lin}, \citenamefont {Gimelshein}, \citenamefont {Antiga} \emph
  {et~al.}}]{PGMLBCKLGA:2019}%
  \BibitemOpen
  \bibfield  {author} {\bibinfo {author} {\bibfnamefont {A.}~\bibnamefont
  {Paszke}}, \bibinfo {author} {\bibfnamefont {S.}~\bibnamefont {Gross}},
  \bibinfo {author} {\bibfnamefont {F.}~\bibnamefont {Massa}}, \bibinfo
  {author} {\bibfnamefont {A.}~\bibnamefont {Lerer}}, \bibinfo {author}
  {\bibfnamefont {J.}~\bibnamefont {Bradbury}}, \bibinfo {author}
  {\bibfnamefont {G.}~\bibnamefont {Chanan}}, \bibinfo {author} {\bibfnamefont
  {T.}~\bibnamefont {Killeen}}, \bibinfo {author} {\bibfnamefont
  {Z.}~\bibnamefont {Lin}}, \bibinfo {author} {\bibfnamefont {N.}~\bibnamefont
  {Gimelshein}}, \bibinfo {author} {\bibfnamefont {L.}~\bibnamefont {Antiga}},
  \emph {et~al.},\ }\bibfield  {title} {\bibinfo {title} {Pytorch: An
  imperative style, high-performance deep learning library},\ }in\ \href@noop
  {} {\emph {\bibinfo {booktitle} {NIPS-W}}}\ (\bibinfo {year} {2019})\ pp.\
  \bibinfo {pages} {8024--8035}\BibitemShut {NoStop}%
\bibitem [{SI()}]{SI}%
  \BibitemOpen
  \href@noop {} {\bibinfo  {journal} {See Supplemental Material at [URL will be
  inserted by publisher] for additional numerical results and specific details
  of the characterization method. However, the article can be fully understood
  without the Supplemental Material}\ }\BibitemShut {NoStop}%
\bibitem [{\citenamefont {Long}\ \emph {et~al.}(2015)\citenamefont {Long},
  \citenamefont {Shelhamer},\ and\ \citenamefont {Darrell}}]{LSD:2015}%
  \BibitemOpen
\bibfield  {journal} {  }\bibfield  {author} {\bibinfo {author} {\bibfnamefont
  {J.}~\bibnamefont {Long}}, \bibinfo {author} {\bibfnamefont {E.}~\bibnamefont
  {Shelhamer}},\ and\ \bibinfo {author} {\bibfnamefont {T.}~\bibnamefont
  {Darrell}},\ }\bibfield  {title} {\bibinfo {title} {Fully convolutional
  networks for semantic segmentation},\ }in\ \href@noop {} {\emph {\bibinfo
  {booktitle} {CVPR}}}\ (\bibinfo {year} {2015})\ pp.\ \bibinfo {pages}
  {3431--3440}\BibitemShut {NoStop}%
\bibitem [{\citenamefont {Ronneberger}\ \emph {et~al.}(2015)\citenamefont
  {Ronneberger}, \citenamefont {Fischer},\ and\ \citenamefont
  {Brox}}]{RFB:2015}%
  \BibitemOpen
  \bibfield  {author} {\bibinfo {author} {\bibfnamefont {O.}~\bibnamefont
  {Ronneberger}}, \bibinfo {author} {\bibfnamefont {P.}~\bibnamefont
  {Fischer}},\ and\ \bibinfo {author} {\bibfnamefont {T.}~\bibnamefont
  {Brox}},\ }\bibfield  {title} {\bibinfo {title} {U-net: Convolutional
  networks for biomedical image segmentation},\ }in\ \href@noop {} {\emph
  {\bibinfo {booktitle} {Med. Image. Comput. Comput. Assist. Interv.}}}\
  (\bibinfo {organization} {Springer},\ \bibinfo {year} {2015})\ pp.\ \bibinfo
  {pages} {234--241}\BibitemShut {NoStop}%
\bibitem [{\citenamefont {Jaeger}(2001)}]{Jaeger:2001}%
  \BibitemOpen
  \bibfield  {author} {\bibinfo {author} {\bibfnamefont {H.}~\bibnamefont
  {Jaeger}},\ }\bibfield  {title} {\bibinfo {title} {The “echo state”
  approach to analysing and training recurrent neural networks-with an erratum
  note},\ }\href@noop {} {\bibfield  {journal} {\bibinfo  {journal} {Bonn,
  Germany: German National Research Center for Information Technology GMD
  Technical Report}\ }\textbf {\bibinfo {volume} {148}},\ \bibinfo {pages} {13}
  (\bibinfo {year} {2001})}\BibitemShut {NoStop}%
\bibitem [{\citenamefont {Jaeger}\ and\ \citenamefont {Haas}(2004)}]{JH:2004}%
  \BibitemOpen
  \bibfield  {author} {\bibinfo {author} {\bibfnamefont {H.}~\bibnamefont
  {Jaeger}}\ and\ \bibinfo {author} {\bibfnamefont {H.}~\bibnamefont {Haas}},\
  }\bibfield  {title} {\bibinfo {title} {Harnessing nonlinearity: Predicting
  chaotic systems and saving energy in wireless communication},\ }\href@noop {}
  {\bibfield  {journal} {\bibinfo  {journal} {Science}\ }\textbf {\bibinfo
  {volume} {304}},\ \bibinfo {pages} {78} (\bibinfo {year} {2004})}\BibitemShut
  {NoStop}%
\bibitem [{\citenamefont {Haynes}\ \emph {et~al.}(2015)\citenamefont {Haynes},
  \citenamefont {Soriano}, \citenamefont {Rosin}, \citenamefont {Fischer},\
  and\ \citenamefont {Gauthier}}]{HSRFG:2015}%
  \BibitemOpen
  \bibfield  {author} {\bibinfo {author} {\bibfnamefont {N.~D.}\ \bibnamefont
  {Haynes}}, \bibinfo {author} {\bibfnamefont {M.~C.}\ \bibnamefont {Soriano}},
  \bibinfo {author} {\bibfnamefont {D.~P.}\ \bibnamefont {Rosin}}, \bibinfo
  {author} {\bibfnamefont {I.}~\bibnamefont {Fischer}},\ and\ \bibinfo {author}
  {\bibfnamefont {D.~J.}\ \bibnamefont {Gauthier}},\ }\bibfield  {title}
  {\bibinfo {title} {Reservoir computing with a single time-delay autonomous
  boolean node},\ }\href {https://doi.org/10.1103/PhysRevE.91.020801}
  {\bibfield  {journal} {\bibinfo  {journal} {Phys. Rev. E}\ }\textbf {\bibinfo
  {volume} {91}},\ \bibinfo {pages} {020801} (\bibinfo {year}
  {2015})}\BibitemShut {NoStop}%
\bibitem [{\citenamefont {Larger}\ \emph {et~al.}(2017)\citenamefont {Larger},
  \citenamefont {Bayl\'on-Fuentes}, \citenamefont {Martinenghi}, \citenamefont
  {Udaltsov}, \citenamefont {Chembo},\ and\ \citenamefont
  {Jacquot}}]{LBMUCJ:2017}%
  \BibitemOpen
  \bibfield  {author} {\bibinfo {author} {\bibfnamefont {L.}~\bibnamefont
  {Larger}}, \bibinfo {author} {\bibfnamefont {A.}~\bibnamefont
  {Bayl\'on-Fuentes}}, \bibinfo {author} {\bibfnamefont {R.}~\bibnamefont
  {Martinenghi}}, \bibinfo {author} {\bibfnamefont {V.~S.}\ \bibnamefont
  {Udaltsov}}, \bibinfo {author} {\bibfnamefont {Y.~K.}\ \bibnamefont
  {Chembo}},\ and\ \bibinfo {author} {\bibfnamefont {M.}~\bibnamefont
  {Jacquot}},\ }\bibfield  {title} {\bibinfo {title} {High-speed photonic
  reservoir computing using a time-delay-based architecture: Million words per
  second classification},\ }\href {https://doi.org/10.1103/PhysRevX.7.011015}
  {\bibfield  {journal} {\bibinfo  {journal} {Phys. Rev. X}\ }\textbf {\bibinfo
  {volume} {7}},\ \bibinfo {pages} {011015} (\bibinfo {year}
  {2017})}\BibitemShut {NoStop}%
\bibitem [{\citenamefont {Pathak}\ \emph {et~al.}(2018)\citenamefont {Pathak},
  \citenamefont {Hunt}, \citenamefont {Girvan}, \citenamefont {Lu},\ and\
  \citenamefont {Ott}}]{PHGLO:2018}%
  \BibitemOpen
  \bibfield  {author} {\bibinfo {author} {\bibfnamefont {J.}~\bibnamefont
  {Pathak}}, \bibinfo {author} {\bibfnamefont {B.}~\bibnamefont {Hunt}},
  \bibinfo {author} {\bibfnamefont {M.}~\bibnamefont {Girvan}}, \bibinfo
  {author} {\bibfnamefont {Z.}~\bibnamefont {Lu}},\ and\ \bibinfo {author}
  {\bibfnamefont {E.}~\bibnamefont {Ott}},\ }\bibfield  {title} {\bibinfo
  {title} {Model-free prediction of large spatiotemporally chaotic systems from
  data: A reservoir computing approach},\ }\href@noop {} {\bibfield  {journal}
  {\bibinfo  {journal} {Phys. Rev. Lett.}\ }\textbf {\bibinfo {volume} {120}},\
  \bibinfo {pages} {024102} (\bibinfo {year} {2018})}\BibitemShut {NoStop}%
\bibitem [{\citenamefont {Carroll}(2018)}]{Carroll:2018}%
  \BibitemOpen
  \bibfield  {author} {\bibinfo {author} {\bibfnamefont {T.~L.}\ \bibnamefont
  {Carroll}},\ }\bibfield  {title} {\bibinfo {title} {Using reservoir computers
  to distinguish chaotic signals},\ }\href
  {https://doi.org/10.1103/PhysRevE.98.052209} {\bibfield  {journal} {\bibinfo
  {journal} {Phys. Rev. E}\ }\textbf {\bibinfo {volume} {98}},\ \bibinfo
  {pages} {052209} (\bibinfo {year} {2018})}\BibitemShut {NoStop}%
\bibitem [{\citenamefont {Nakai}\ and\ \citenamefont {Saiki}(2018)}]{NS:2018}%
  \BibitemOpen
  \bibfield  {author} {\bibinfo {author} {\bibfnamefont {K.}~\bibnamefont
  {Nakai}}\ and\ \bibinfo {author} {\bibfnamefont {Y.}~\bibnamefont {Saiki}},\
  }\bibfield  {title} {\bibinfo {title} {Machine-learning inference of fluid
  variables from data using reservoir computing},\ }\href
  {https://doi.org/10.1103/PhysRevE.98.023111} {\bibfield  {journal} {\bibinfo
  {journal} {Phys. Rev. E}\ }\textbf {\bibinfo {volume} {98}},\ \bibinfo
  {pages} {023111} (\bibinfo {year} {2018})}\BibitemShut {NoStop}%
\bibitem [{\citenamefont {Roland}\ and\ \citenamefont
  {Parlitz}(2018)}]{ZP:2018}%
  \BibitemOpen
  \bibfield  {author} {\bibinfo {author} {\bibfnamefont {Z.~S.}\ \bibnamefont
  {Roland}}\ and\ \bibinfo {author} {\bibfnamefont {U.}~\bibnamefont
  {Parlitz}},\ }\bibfield  {title} {\bibinfo {title} {Observing spatio-temporal
  dynamics of excitable media using reservoir computing},\ }\href@noop {}
  {\bibfield  {journal} {\bibinfo  {journal} {Chaos}\ }\textbf {\bibinfo
  {volume} {28}},\ \bibinfo {pages} {043118} (\bibinfo {year}
  {2018})}\BibitemShut {NoStop}%
\bibitem [{\citenamefont {Weng}\ \emph {et~al.}(2019)\citenamefont {Weng},
  \citenamefont {Yang}, \citenamefont {Gu}, \citenamefont {Zhang},\ and\
  \citenamefont {Small}}]{WYGZS:2019}%
  \BibitemOpen
  \bibfield  {author} {\bibinfo {author} {\bibfnamefont {T.}~\bibnamefont
  {Weng}}, \bibinfo {author} {\bibfnamefont {H.}~\bibnamefont {Yang}}, \bibinfo
  {author} {\bibfnamefont {C.}~\bibnamefont {Gu}}, \bibinfo {author}
  {\bibfnamefont {J.}~\bibnamefont {Zhang}},\ and\ \bibinfo {author}
  {\bibfnamefont {M.}~\bibnamefont {Small}},\ }\bibfield  {title} {\bibinfo
  {title} {Synchronization of chaotic systems and their machine-learning
  models},\ }\href {https://doi.org/10.1103/PhysRevE.99.042203} {\bibfield
  {journal} {\bibinfo  {journal} {Phys. Rev. E}\ }\textbf {\bibinfo {volume}
  {99}},\ \bibinfo {pages} {042203} (\bibinfo {year} {2019})}\BibitemShut
  {NoStop}%
\bibitem [{\citenamefont {Jiang}\ and\ \citenamefont {Lai}(2019)}]{JL:2019}%
  \BibitemOpen
  \bibfield  {author} {\bibinfo {author} {\bibfnamefont {J.}~\bibnamefont
  {Jiang}}\ and\ \bibinfo {author} {\bibfnamefont {Y.-C.}\ \bibnamefont
  {Lai}},\ }\bibfield  {title} {\bibinfo {title} {Model-free prediction of
  spatiotemporal dynamical systems with recurrent neural networks: Role of
  network spectral radius},\ }\href@noop {} {\bibfield  {journal} {\bibinfo
  {journal} {Phys. Rev. Res.}\ }\textbf {\bibinfo {volume} {1}},\ \bibinfo
  {pages} {033056} (\bibinfo {year} {2019})}\BibitemShut {NoStop}%
\bibitem [{\citenamefont {Vlachas}\ \emph {et~al.}(2020)\citenamefont
  {Vlachas}, \citenamefont {Pathak}, \citenamefont {Hunt}, \citenamefont
  {Sapsis}, \citenamefont {Girvan}, \citenamefont {Ott},\ and\ \citenamefont
  {Koumoutsakos}}]{VPHSGOK:2020}%
  \BibitemOpen
  \bibfield  {author} {\bibinfo {author} {\bibfnamefont {P.~R.}\ \bibnamefont
  {Vlachas}}, \bibinfo {author} {\bibfnamefont {J.}~\bibnamefont {Pathak}},
  \bibinfo {author} {\bibfnamefont {B.~R.}\ \bibnamefont {Hunt}}, \bibinfo
  {author} {\bibfnamefont {T.~P.}\ \bibnamefont {Sapsis}}, \bibinfo {author}
  {\bibfnamefont {M.}~\bibnamefont {Girvan}}, \bibinfo {author} {\bibfnamefont
  {E.}~\bibnamefont {Ott}},\ and\ \bibinfo {author} {\bibfnamefont
  {P.}~\bibnamefont {Koumoutsakos}},\ }\bibfield  {title} {\bibinfo {title}
  {Backpropagation algorithms and reservoir computing in recurrent neural
  networks for the forecasting of complex spatiotemporal dynamics},\
  }\href@noop {} {\bibfield  {journal} {\bibinfo  {journal} {Neural Net.}\
  }\textbf {\bibinfo {volume} {126}},\ \bibinfo {pages} {191} (\bibinfo {year}
  {2020})}\BibitemShut {NoStop}%
\bibitem [{\citenamefont {Fan}\ \emph {et~al.}(2020)\citenamefont {Fan},
  \citenamefont {Jiang}, \citenamefont {Zhang}, \citenamefont {Wang},\ and\
  \citenamefont {Lai}}]{FJZWL:2020}%
  \BibitemOpen
  \bibfield  {author} {\bibinfo {author} {\bibfnamefont {H.}~\bibnamefont
  {Fan}}, \bibinfo {author} {\bibfnamefont {J.}~\bibnamefont {Jiang}}, \bibinfo
  {author} {\bibfnamefont {C.}~\bibnamefont {Zhang}}, \bibinfo {author}
  {\bibfnamefont {X.}~\bibnamefont {Wang}},\ and\ \bibinfo {author}
  {\bibfnamefont {Y.-C.}\ \bibnamefont {Lai}},\ }\bibfield  {title} {\bibinfo
  {title} {Long-term prediction of chaotic systems with machine learning},\
  }\href {https://doi.org/10.1103/PhysRevResearch.2.012080} {\bibfield
  {journal} {\bibinfo  {journal} {Phys. Rev. Research}\ }\textbf {\bibinfo
  {volume} {2}},\ \bibinfo {pages} {012080} (\bibinfo {year}
  {2020})}\BibitemShut {NoStop}%
\bibitem [{\citenamefont {Kong}\ \emph {et~al.}(2021)\citenamefont {Kong},
  \citenamefont {Fan}, \citenamefont {Grebogi},\ and\ \citenamefont
  {Lai}}]{KFGL:2021a}%
  \BibitemOpen
  \bibfield  {author} {\bibinfo {author} {\bibfnamefont {L.-W.}\ \bibnamefont
  {Kong}}, \bibinfo {author} {\bibfnamefont {H.-W.}\ \bibnamefont {Fan}},
  \bibinfo {author} {\bibfnamefont {C.}~\bibnamefont {Grebogi}},\ and\ \bibinfo
  {author} {\bibfnamefont {Y.-C.}\ \bibnamefont {Lai}},\ }\bibfield  {title}
  {\bibinfo {title} {Machine learning prediction of critical transition and
  system collapse},\ }\href {https://doi.org/10.1103/PhysRevResearch.3.013090}
  {\bibfield  {journal} {\bibinfo  {journal} {Phys. Rev. Research}\ }\textbf
  {\bibinfo {volume} {3}},\ \bibinfo {pages} {013090} (\bibinfo {year}
  {2021})}\BibitemShut {NoStop}%
\bibitem [{\citenamefont {Cox}\ and\ \citenamefont {Matthews}(2002)}]{CM:2002}%
  \BibitemOpen
  \bibfield  {author} {\bibinfo {author} {\bibfnamefont {S.~M.}\ \bibnamefont
  {Cox}}\ and\ \bibinfo {author} {\bibfnamefont {P.~C.}\ \bibnamefont
  {Matthews}},\ }\bibfield  {title} {\bibinfo {title} {Exponential time
  differencing for stiff systems},\ }\href@noop {} {\bibfield  {journal}
  {\bibinfo  {journal} {J. Comput. Phys.}\ }\textbf {\bibinfo {volume} {176}},\
  \bibinfo {pages} {430} (\bibinfo {year} {2002})}\BibitemShut {NoStop}%
\bibitem [{\citenamefont {Dee}\ \emph {et~al.}(2011)\citenamefont {Dee},
  \citenamefont {Uppala}, \citenamefont {Simmons}, \citenamefont {Berrisford},
  \citenamefont {Poli}, \citenamefont {Kobayashi}, \citenamefont {Andrae},
  \citenamefont {Balmaseda}, \citenamefont {Balsamo}, \citenamefont {Bauer}
  \emph {et~al.}}]{DUSBPKABBB:2011}%
  \BibitemOpen
  \bibfield  {author} {\bibinfo {author} {\bibfnamefont {D.~P.}\ \bibnamefont
  {Dee}}, \bibinfo {author} {\bibfnamefont {S.}~\bibnamefont {Uppala}},
  \bibinfo {author} {\bibfnamefont {A.}~\bibnamefont {Simmons}}, \bibinfo
  {author} {\bibfnamefont {P.}~\bibnamefont {Berrisford}}, \bibinfo {author}
  {\bibfnamefont {P.}~\bibnamefont {Poli}}, \bibinfo {author} {\bibfnamefont
  {S.}~\bibnamefont {Kobayashi}}, \bibinfo {author} {\bibfnamefont
  {U.}~\bibnamefont {Andrae}}, \bibinfo {author} {\bibfnamefont
  {M.}~\bibnamefont {Balmaseda}}, \bibinfo {author} {\bibfnamefont
  {G.}~\bibnamefont {Balsamo}}, \bibinfo {author} {\bibfnamefont {d.~P.}\
  \bibnamefont {Bauer}}, \emph {et~al.},\ }\bibfield  {title} {\bibinfo {title}
  {The era-interim reanalysis: Configuration and performance of the data
  assimilation system},\ }\href@noop {} {\bibfield  {journal} {\bibinfo
  {journal} {Q. J. Roy. Meteor. Soc.}\ }\textbf {\bibinfo {volume} {137}},\
  \bibinfo {pages} {553} (\bibinfo {year} {2011})}\BibitemShut {NoStop}%
\bibitem [{\citenamefont {Singleton}(2008)}]{Singleton:2008}%
  \BibitemOpen
  \bibfield  {author} {\bibinfo {author} {\bibfnamefont {F.}~\bibnamefont
  {Singleton}},\ }\bibfield  {title} {\bibinfo {title} {The beaufort scale of
  winds--its relevance, and its use by sailors},\ }\href@noop {} {\bibfield
  {journal} {\bibinfo  {journal} {Weather}\ }\textbf {\bibinfo {volume} {63}},\
  \bibinfo {pages} {37} (\bibinfo {year} {2008})}\BibitemShut {NoStop}%
\end{thebibliography}

%
\end{document}